%% file: main.tex
\documentclass[11pt]{article}
\usepackage[final]{acl}
\usepackage{times}
\usepackage{latexsym}
\usepackage[T1]{fontenc}
\usepackage[utf8]{inputenc}
\usepackage{microtype}
\usepackage{inconsolata}
\usepackage{graphicx}
\usepackage{booktabs}
\usepackage{xcolor}
\usepackage{listings}
\usepackage{enumitem}
\usepackage{amsmath}
\usepackage{amssymb}
\usepackage[most]{tcolorbox}

\definecolor{darkblue}{rgb}{0, 0, 0.5}
\hypersetup{
  colorlinks=true,
  citecolor=darkblue,
  linkcolor=darkblue,
  urlcolor=darkblue,
  pdftitle={BAGEL: Benchmarking Animal knowledGe Expertise in Language Models},
  pdfauthor={Jiacheng Shen, Masato Hagiwara, Milad Alizadeh, Ellen Gilsenan-McMahon, Marius Miron, David Robinson, Emmanuel Chemla, Sara Keen, Gagan Narula, Mathieu Lauriere, Matthieu Geist, Olivier Pietquin}
}
\definecolor{promptbg}{RGB}{245,247,250}
\definecolor{promptframe}{RGB}{180,190,205}
\definecolor{prompttitle}{RGB}{60,72,88}

\lstdefinestyle{promptstyle}{
    basicstyle=\ttfamily\footnotesize,
    breaklines=true,
    columns=fullflexible,
    keepspaces=true,
    showstringspaces=false,
    frame=none,
    xleftmargin=0pt,
    xrightmargin=0pt,
    aboveskip=0pt,
    belowskip=0pt
}

\newtcblisting{promptbox}[2][]{
    enhanced,
    breakable,
    colback=promptbg,
    colframe=promptframe,
    coltitle=prompttitle,
    boxrule=0.6pt,
    arc=2mm,
    left=1.2mm,
    right=1.2mm,
    top=1mm,
    bottom=1mm,
    listing only,
    listing options={style=promptstyle},
    title={#2},
    fonttitle=\bfseries,
    #1
}

\newtcolorbox{examplebox}[2][]{
    enhanced,
    colback=white,
    colframe=black!40,
    boxrule=0.5pt,
    arc=1mm,
    left=1.2mm,
    right=1.2mm,
    top=1mm,
    bottom=1mm,
    title={#2},
    fonttitle=\bfseries,
    #1
}

\title{BAGEL: \textbf{B}enchmarking \textbf{A}nimal knowled\textbf{G}e \textbf{E}xpertise in \textbf{L}anguage Models}

\author{%
  \textbf{Jiacheng Shen\textsuperscript{2$\dagger$}},
  \textbf{Masato Hagiwara\textsuperscript{1$\dagger$}},
  \textbf{Milad Alizadeh\textsuperscript{1}},
  \textbf{Ellen Gilsenan-McMahon\textsuperscript{1}}\\
  \textbf{Marius Miron\textsuperscript{1}},
  \textbf{David Robinson\textsuperscript{1}},
  \textbf{Emmanuel Chemla\textsuperscript{1}},
  \textbf{Sara Keen\textsuperscript{1}}\\
  \textbf{Gagan Narula\textsuperscript{1}},
  \textbf{Mathieu Lauri\`ere\textsuperscript{3$\ddagger$}},
  \textbf{Matthieu Geist\textsuperscript{1$\ddagger$}},
  \textbf{Olivier Pietquin\textsuperscript{1$\ddagger$}}\\[0.35em]
  \textsuperscript{1}Earth Species Project\\
  \textsuperscript{2}NYU Center for Data Science; NYU Shanghai\\
  \textsuperscript{3}NYU Shanghai Center for Data Science;\\
  NYU-ECNU Institute of Mathematical Sciences at NYU Shanghai\\
  \textsuperscript{$\dagger$}Equal contribution. \quad
  \textsuperscript{$\ddagger$}Co-supervising authors.
}

\begin{document}
\maketitle
\begin{abstract}
Large language models perform well on broad knowledge benchmarks, yet their animal-domain competence remains poorly characterized under a consistent protocol. We introduce BAGEL (\textbf{B}enchmarking \textbf{A}nimal knowled\textbf{G}e \textbf{E}xpertise in \textbf{L}anguage models), a 6,822-question multiple-choice benchmark drawn from Wikipedia, Global Biotic Interactions (GloBI), and bioRxiv. It spans encyclopedic species knowledge, ecological interaction inference, and interpretation of animal-focused preprints. Questions are generated automatically and screened by GPT-5.5 using a rubric calibrated on 150 expert-reviewed items; this process is not exhaustive human adjudication. We evaluate 14 models without external retrieval, using model parameters and question-local evidence. Accuracy varies sharply by source and question type: frontier models that excel on Wikipedia and bioRxiv remain comparatively weak on GloBI participant identification, where pooled errors reveal ecological role confusion. A 100-item post-release bioRxiv probe provides preliminary evidence that GPT-5.4's literature performance is not entirely due to verbatim recall, without ruling out broader training exposure. Proprietary checkpoints lead the best open-weight systems overall, but aggregate scores incompletely characterize animal-domain competence.
\end{abstract}
\section{Introduction}

Instruction-tuned language models have improved rapidly on general knowledge and reasoning benchmarks~\citep{ouyang2022instructgpt,openai2023gpt4}, including suites such as MMLU~\citep{hendrycks2021mmlu} and ScienceQA~\citep{lu2022learn}. These gains have encouraged their use for scientific information access, literature synthesis, and downstream tool building. Strong performance on broad tests does not, however, establish whether a model can reliably answer specialized questions about the natural world---for example, species-level taxonomy, ecological relationships, or natural-history reasoning grounded in organismal biology. Animal knowledge is heterogeneous: an encyclopedic fact about morphology, a directed host--parasite link, and a conclusion drawn from a preprint abstract require different kinds of evidence and inference, yet practitioners increasingly rely on the same text backbone for all of them.

This gap is becoming practically important. Ecologists have used LLMs to extract structured information from literature, including host--pathogen records and species interactions at scale~\citep{gougherty2024ecological,keck2025massive}, and recent evaluations report uneven ecological competence across task types~\citep{dorm2025ecologicalknowledge}. In bioacoustics, resources such as BEANS~\citep{hagiwara2022beans}, ISPA~\citep{hagiwara2024ispa}, and NatureLM-audio~\citep{robinson2024naturelmaudio} target animal sounds and audio--language modeling; BioCLIP extends similar ideas to visual recognition across the tree of life~\citep{stevens2024bioclip}. Much of this work emphasizes extraction pipelines, audio understanding, or vision rather than asking, under one fixed protocol, how much animal-related knowledge a text-only model already encodes. Without such a baseline, it is difficult to decide whether a biodiversity application should trust parametric knowledge, retrieve from external databases, or combine both.

We introduce BAGEL (\textbf{B}enchmarking \textbf{A}nimal knowled\textbf{G}e \textbf{E}xpertise in \textbf{L}anguage models), a multiple-choice benchmark for closed-book evaluation of animal knowledge and reasoning.\footnote{Dataset release: \url{https://huggingface.co/datasets/EarthSpeciesProject/BAGEL}.} The benchmark comprises 6,822 validated questions drawn from Wikipedia, Global Biotic Interactions (GloBI)~\citep{poelen2014globi}, and bioRxiv. Wikipedia probes encyclopedic species knowledge across eight natural-history dimensions; GloBI probes ecological interaction reasoning from documented species links; bioRxiv probes interpretation of animal-focused preprints. We do not argue that deployed biodiversity systems should operate without retrieval---in practice they almost always combine models with databases, ontologies, and sensors~\citep{lewis2020rag,mallen2023nottrust}. BAGEL instead provides a controlled diagnostic of what models can answer without external retrieval, using their parameters and any evidence contained in the question stem. This distinction matters for bioRxiv, where self-contained stems test interpretation of reported findings rather than recall of the source document. The same baseline is useful when selecting text backbones for multimodal systems such as NatureLM-audio~\citep{robinson2024naturelmaudio} and BioCLIP~\citep{stevens2024bioclip}, and for future agentic approaches that decide when to call biodiversity databases or tools.

BAGEL is built automatically from public sources and audited before release (Section~\ref{sec:human_validation}), yielding the 6,822-question dataset evaluated in this paper. We evaluate 14 models spanning frontier API checkpoints and open-weight systems from 360M to 32B parameters, report accuracy by source and question type, and analyze pooled failure patterns. Three findings stand out. First, accuracy varies sharply across sources: even strong models that excel on Wikipedia and bioRxiv remain comparatively weak on GloBI interaction items. Second, within GloBI, interaction-type inference is much easier than participant identification, so headline domain scores average subtasks of very different difficulty. Third, pooled error analysis reveals recurring ecological role confusion rather than idiosyncratic noise, while only a small set of items defeats every evaluated model. BAGEL therefore functions less as a single leaderboard and more as a structured map of where text models are already informative about animals and where they are not.

The remainder of the paper is organized as follows. Section~\ref{sec:benchmark_construction} describes benchmark construction and validation. Section~\ref{sec:experimental_setup} details the evaluation protocol. Section~\ref{sec:results_discussion} presents main results, per-question-type breakdowns, error analysis, and a post-release bioRxiv probe. Section~\ref{sec:conclusion} concludes; limitations are discussed in a dedicated section after the conclusion. The appendix provides prompts, extended tables, and additional analyses.
\looseness=-1
\section{Related Work}

Work on AI for biodiversity spans ecology, bioacoustics, and vision, but evaluation protocols remain fragmented. LLMs have been used to extract ecological variables from disease reports~\citep{gougherty2024ecological} and species interactions from large corpora~\citep{keck2025massive}; direct tests of ecological knowledge find strong factual recall paired with weaker reasoning~\citep{dorm2025ecologicalknowledge}. Bioacoustic benchmarks and models---including BEANS~\citep{hagiwara2022beans}, ISPA~\citep{hagiwara2024ispa}, and NatureLM-audio~\citep{robinson2024naturelmaudio}---focus on sound and audio--language tasks, while BioCLIP~\citep{stevens2024bioclip} targets visual recognition across taxa. Specialized geoscience and ocean language models~\citep{deng2023k2,bi2023oceangpt} point in a similar direction for other environmental domains. These lines of work demonstrate growing interest in foundation models for nature, yet they rarely evaluate whether a text-only model can answer animal questions \emph{without} access to the source documents used to construct those questions.

General knowledge and science benchmarks provide useful baselines but do not emphasize fine-grained natural-history content. MMLU~\citep{hendrycks2021mmlu} remains the standard broad knowledge suite, and ScienceQA~\citep{lu2022learn} adds multimodal school-science questions. Biomedical suites such as BLURB~\citep{gu2021domain}, PubMedQA~\citep{jin2019pubmedqa}, and BioASQ~\citep{nentidis2023bioasq} show the value of domain-specific evaluation, but their scope is human health and biomedical text rather than organismal diversity. More recent environmental benchmarks, including EnviroExam~\citep{huang2024enviroexam}, ELLE~\citep{guo2025elle}, and ecological evaluations of LLM competence~\citep{dorm2025ecologicalknowledge}, broaden coverage of sustainability- and ecology-themed tasks. BAGEL complements this landscape with a narrower focus on animal natural history and a single closed-book format across three public text sources, with reporting by domain and question type rather than one pooled academic score. Appendix Table~\ref{tab:related_benchmarks} situates BAGEL among neighboring benchmarks; Appendix~\ref{app:mmlu_vs_bagel} further shows that BAGEL correlates with MMLU-Pro but does not preserve model rankings.

\begin{figure*}[t]
\centering
\includegraphics[width=\linewidth]{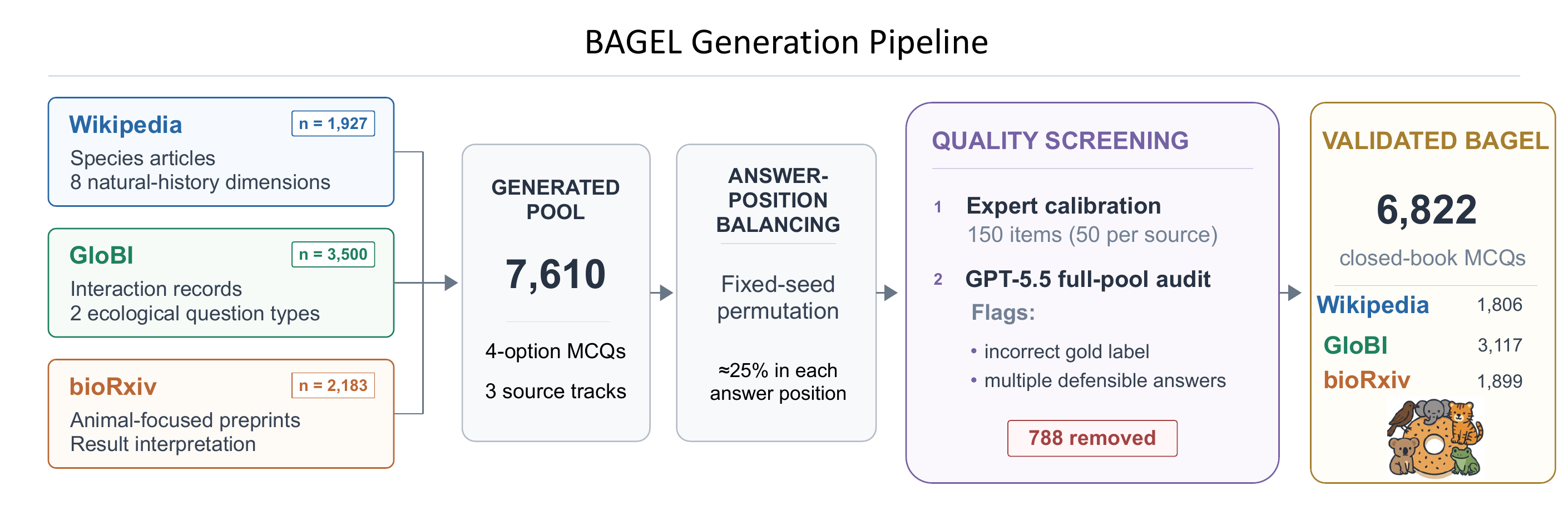}
\caption{Overview of the BAGEL curation pipeline. GPT-4o-mini generates 7{,}610 questions from three public sources. After answer-position balancing, an expert-calibrated GPT-5.5 audit removes 788 items with incorrect gold labels or multiple defensible answers, yielding 6{,}822 validated closed-book questions.}
\label{fig:bagel_bench_curation}
\end{figure*}
\section{Benchmark Construction}
\label{sec:benchmark_construction}
Figure~\ref{fig:bagel_bench_curation} outlines the curation pipeline. The three corpora are public, machine-readable, and complementary: encyclopedic species facts from Wikipedia, ecological interactions from GloBI, and animal-focused preprints from bioRxiv. We froze the construction pools between February and March 2026. They do not exhaust natural-history competence, but together they support a structured evaluation of text-only animal knowledge.

\subsection{Wikipedia}
\label{subsec:benchmark_wikipedia}
The Wikipedia track evaluates encyclopedic knowledge about animals. The fixed candidate inventory contains 263 chordate taxa: 146 birds, 95 mammals, and 22 amphibians. English Wikipedia species articles supply the evidence used during construction; at test time the model sees only the question and four answer options, not the article.\footnote{Portal \url{https://www.wikipedia.org/}; taxa are linked through Wikidata, and article plain-text extracts are retrieved via the MediaWiki API.}

For each taxon we resolve the English Wikipedia title via Wikidata and fetch the plain-text extract through the MediaWiki API, skipping pages with missing text or fewer than 1{,}000 characters to avoid stubs. Extracts longer than 180{,}000 characters are truncated at a paragraph or sentence boundary so prompts fit the generator's context window. GPT-4o-mini then produces up to eight four-option items per species (Appendix~\ref{wiki_bench_gen_prompt}), each tagged with one of eight dimensions: Taxonomy, Behavior, Communication, Morphology, Habitat, Cognition, Geographic Distribution, or Diet. Items must be supported by the article text; unsupported dimensions are skipped. We retain outputs that pass basic format checks (valid dimension, four options, one marked correct answer).

These dimensions reflect recurring fields in species accounts and natural-history resources, including taxonomy, morphology, distribution, habitat, behavior, communication, and feeding ecology. GloBI and bioRxiv complement these encyclopedic dimensions with interaction inference and scientific-result interpretation, respectively.

\subsection{Global Biotic Interactions (GloBI)}
GloBI records describe directed links between taxa, with an interaction label and optional locality, date, coordinates, and provenance.\footnote{GloBI portal: \url{https://www.globalbioticinteractions.org/}.}

We read the first 10{,}000 rows of a March 2026 GloBI stable-download table and normalize field names across common export formats. Rows without both endpoint taxa and an interaction-type label are excluded. Each retained row is converted into a short natural-language summary of the interaction together with optional locality, date, and coordinate metadata, plus dataset- and reference-level provenance. We keep 3{,}500 interactions, stratifying by interaction type so relation labels are more evenly represented than in a random row sample. When types tie, we prefer rows with richer metadata (locality, coordinates, dates), using seed~42; the final list is shuffled before question generation.

GPT-4o-mini receives only the interaction summary and must return one four-option item of type Masked participant identification or Masked interaction type inference (Appendix~\ref{app:globi_bench_gen_prompt}). The prompt discourages giveaway verbs and asks for ecologically plausible distractors. Unparseable outputs are dropped.

\subsection{bioRxiv}
\label{subsec:benchmark_biorxiv}
The bioRxiv track tests whether models can reason about animal research described in preprints. We collect posts published from January through December 2023 in Animal Behavior and Cognition, Ecology, Evolutionary Biology, and Zoology\footnote{bioRxiv: \url{https://www.biorxiv.org/}.} via the site's advanced search. We concatenate title, abstract, and body, normalize whitespace, drop duplicate paragraphs, and remove paragraphs that start with ``Figure''. GPT-4o-mini reads each document and writes one Result Interpretation item (Appendix~\ref{app:biorxiv_bench_gen_prompt}): the stem must include enough setup and findings to answer without the PDF. We also evaluate GPT-5.4 on a separate post-release holdout described in Section~\ref{subsec:fresh_biorxiv}.

\subsection{Answer-position balancing}
Before option shuffling, the correct option was heavily skewed toward particular positions within each domain (Appendix~\ref{app:answer_position_bias}, Table~\ref{tab:appendix-answer-position-dataset}). Because models are sensitive to option order in multiple-choice tests~\citep{zheng2023llms,pezeshkpour2024large}, we randomly permute the four choices for every item with a fixed seed and update the gold label accordingly. After shuffling, each position accounts for roughly 25\% of correct answers (Table~\ref{tab:appendix-answer-position-after-shuffle}).

\subsection{Human validation}
\label{sec:human_validation}
All items are machine-generated, so we audit quality before release. One author with relevant domain expertise reviewed 50 examples from each of Wikipedia, GloBI, and bioRxiv (150 total), checking gold-answer correctness and whether other options were also defensible. No external annotators were recruited or compensated, and the audit collected no personal data. To stress-test the filter, sampling was stratified between items missed by all eight models in an initial evaluation and standard items. The bioRxiv and GloBI samples each contained 25 items from each stratum; Wikipedia contributed all 20 available all-model failures and 30 standard items. GPT-5.5 applied the same rubric on this challenge-enriched audit set and agreed with the expert on 92\% of the two filter labels used for release (incorrect gold answer; multiple plausible answers). On the corresponding remove-or-keep decision, agreement was 94.7\%, with GPT-5.5 tending to over-flag relative to the expert. We then screened the full pool with GPT-5.5 and removed flagged items. Table~\ref{tab:validation_filter_counts} reports retention by domain. This audit agreement measures judge consistency on the 150-item sample; it is not a residual error rate on the validated release. Unless stated otherwise, all results use the resulting 6{,}822-question dataset. Appendix~\ref{app:human_audit_protocol} reproduces the author-audit instructions and decision rule.

\begin{table}[t]
\centering
\small
\setlength{\tabcolsep}{6pt}
\begin{tabular}{lrrr}
\hline
\textbf{Domain} & \textbf{Generated} & \textbf{Validated} & \textbf{Removed} \\
\hline
Wikipedia & 1,927 & 1,806 & 121 \\
GloBI & 3,500 & 3,117 & 383 \\
bioRxiv & 2,183 & 1,899 & 284 \\
\hline
\textbf{Total} & \textbf{7,610} & \textbf{6,822} & \textbf{788} \\
\hline
\end{tabular}
\caption{Question counts before and after validation. All main-paper results use the validated release ($n{=}6{,}822$).}
\label{tab:validation_filter_counts}
\end{table}

Representative examples from each source appear in Appendix~\ref{app:example_items}.

\section{Experimental Setup}
\label{sec:experimental_setup}

We evaluate GPT-5.4, Claude Opus~4.6, and Gemini 3.1 Flash Lite, together with open-weight models ranging from SmolLM2-360M-Instruct~\citep{benallal2025smollm2} through Qwen3 (0.6B--32B, reasoning disabled)~\citep{yang2025qwen3}, Gemma 3 27B IT and Gemma-7B~\citep{gemmateam2025gemma3technicalreport,gemmateam2024gemma}, Llama~3.1-8B Instruct~\citep{grattafiori2024llama3}, Mistral-7B-Instruct~\citep{jiang2023mistral}, and Phi-4~\citep{abdin2024phi4}. The exact API identifiers are \texttt{gpt-5.4}, \texttt{claude-opus-4-6}, and \texttt{gemini-3.1-flash-lite}; their evaluations were conducted between March~28 and June~9, 2026. Each model answers every question in a closed-book setting: the prompt contains only the instruction, question stem, and four options, with no access to the source text used during construction. We decode greedily with seed~0 and report one run per model. Accuracy is computed per source domain, per question type where applicable, and as an unweighted macro average across the three domains. The evaluation prompt is provided in Appendix~\ref{app:evaluation_prompt}.

BAGEL comprises 6{,}822 four-option questions: 1{,}899 from bioRxiv, 3{,}117 from GloBI, and 1{,}806 from Wikipedia (Table~\ref{tab:validation_filter_counts}). Appendix Table~\ref{tab:bagel_full_stats} breaks these down by dimension. Appendix Table~\ref{tab:bagel_difficulty_three_frontier} assigns each item an easy, medium, or hard label based on whether GPT-5.4, Claude Opus~4.6, and Gemini 3.1 Flash Lite all answer correctly, exactly one or two do, or none does. We treat these strata as a relative difficulty signal induced by frontier-model agreement, not as an independent measure of biological hardness. Mean stem length is 180.5 characters and mean option length is 35.7 characters. Fourteen models have complete predictions on all three domains and are used in the pooled error analysis in Section~\ref{subsec:error_analysis}.

\section{Results and Discussion}
\label{sec:results_discussion}

\subsection{Overall accuracy across sources}
\label{subsec:overall_accuracy}

\begin{table*}[t!]
\centering
\small
\setlength{\tabcolsep}{5pt}
\begin{tabular}{llccc}
\hline
Model  & GloBI & Wiki & bioRxiv & Overall \\
\hline
Empirical Random Guess & 0.2686 & 0.2247 & 0.2652 & 0.2528 \\
\hline
\multicolumn{5}{c}{\textbf{Claude}} \\
\hline
Claude Opus 4.6 & 0.8271 & \underline{0.9413} & 0.9479 & 0.9054 \\
\hline
\multicolumn{5}{c}{\textbf{GPT}} \\
\hline
GPT-5.4  & 0.7857 & 0.9120 & \underline{0.9763} & 0.8913 \\
\hline
\multicolumn{5}{c}{\textbf{Gemini}} \\
\hline
Gemini 3.1 Flash Lite & \underline{0.8293} & 0.9225 & 0.9679 & \underline{0.9066} \\
\hline
\multicolumn{5}{c}{\textbf{Gemma}} \\
\hline
Gemma-7B & 0.4790 & 0.6628 & 0.8847 & 0.6755 \\
Gemma 3 27B IT & \textbf{0.7559} & 0.8173 & 0.9405 & \textbf{0.8379} \\
\hline
\multicolumn{5}{c}{\textbf{Llama}} \\
\hline
Llama 3.1-8B Instruct & 0.6397 & 0.7514 & 0.9142 & 0.7684 \\
\hline
\multicolumn{5}{c}{\textbf{Mistral}} \\
\hline
Mistral-7B-Instruct-v0.3 & 0.6057 & 0.6982 & 0.8799 & 0.7280 \\
\hline
\multicolumn{5}{c}{\textbf{Phi}} \\
\hline
Phi-4 & 0.7353 & 0.8106 & 0.9558 & 0.8339 \\
\hline
\multicolumn{5}{c}{\textbf{Qwen3}} \\
\hline
Qwen3-0.6B & 0.2496 & 0.2458 & 0.2407 & 0.2454 \\
Qwen3-4B & 0.5887 & 0.6916 & 0.9189 & 0.7331 \\
Qwen3-8B & 0.6596 & 0.7769 & 0.9273 & 0.7879 \\
Qwen3-14B & 0.6965 & 0.8062 & 0.9394 & 0.8140 \\
Qwen3-32B & 0.7193 & \textbf{0.8228} & \textbf{0.9531} & 0.8317 \\
\hline
\multicolumn{5}{c}{\textbf{SmolLM2}} \\
\hline
SmolLM2-360M-Instruct & 0.2454 & 0.2381 & 0.2517 & 0.2451 \\
\hline
\end{tabular}
\caption{Accuracy on BAGEL after validation ($n{=}6{,}822$), grouped by model family with sizes ascending within each family. \textbf{Bold} marks the best open-weight score in each column; among proprietary API models, the best score in each column is \underline{underlined}. Overall is the unweighted mean of the three domain accuracies. The empirical random baseline is one fixed-seed draw; its Overall value is computed by the same macro average.}
\label{tab:domain_accuracy}
\end{table*}

Table~\ref{tab:domain_accuracy} reports accuracy on BAGEL. Performance varies substantially across sources: even the strongest models score much higher on Wikipedia and bioRxiv than on GloBI, showing that aggregate animal-domain competence depends strongly on source and question type. Among proprietary API models, Gemini 3.1 Flash Lite has the highest three-domain macro point estimate (0.9066), closely followed by Claude Opus~4.6 (0.9054) and then GPT-5.4 (0.8913). Gemini also leads on GloBI (0.8293), whereas Claude Opus~4.6 remains strongest on Wikipedia and GPT-5.4 on bioRxiv. Among open-weight systems, Gemma~3 27B IT has the highest overall point estimate (0.8379), followed by Phi-4 (0.8339) and Qwen3-32B (0.8317); Llama~3.1-8B is the strongest sub-10B model (0.7684). The smallest open models perform near chance, whereas mid-sized checkpoints already reach reasonable accuracy on Wikipedia and bioRxiv while remaining comparatively weak on GloBI.

Because the evaluation excludes external retrieval and tools~\citep{lewis2020rag,mallen2023nottrust}, scores reflect what models can answer from their parameters and question-local evidence. They are informative when selecting a text backbone but do not prescribe how deployed biodiversity systems should be built. Animal facts are long-tailed, so this controlled pass helps identify where external knowledge would be required; the bioRxiv track additionally measures interpretation of findings stated in the question. Within the Qwen3 family (0.6B--32B), accuracy increases from near-random at 0.6B to roughly 0.73--0.83 for larger checkpoints, with gains on all three domains up to 32B (Appendix Figure~\ref{fig:qwen3_family_curves}). Answer positions are shuffled to reduce letter bias (Section~\ref{sec:benchmark_construction}; Appendix~\ref{app:answer_position_bias}), although some models still show positional preferences in their outputs.

\subsection{Accuracy by question type}
\label{subsec:question_type_accuracy}

Aggregate domain scores can be misleading, especially on GloBI. Table~\ref{tab:globi_breakdown} separates the two GloBI question types: Masked interaction type inference and Masked participant identification. Frontier models reach roughly 0.94--0.96 on interaction-type inference but only 0.56--0.68 on participant identification; GPT-5.4, for example, scores 0.95 on the former and 0.56 on the latter. Open-weight models show the same pattern at lower absolute levels. The headline GloBI column in Table~\ref{tab:domain_accuracy} therefore averages subtasks of very different difficulty. Wikipedia accuracy also varies by dimension---Claude Opus~4.6 exceeds 0.97 on Habitat but only 0.90 on Behavior---although the spread is narrower than on GloBI (Appendix Table~\ref{tab:wiki_breakdown}).

\begin{table*}[t]
\centering
\small
\setlength{\tabcolsep}{8pt}
\begin{tabular}{lcc}
\hline
\textbf{Model} & \textbf{Masked interaction type inference} & \textbf{Masked participant identification} \\
\hline
SmolLM2-360M-Instruct & 0.258 & 0.229 \\
Qwen3-0.6B & 0.249 & 0.250 \\
Qwen3-4B & 0.834 & 0.250 \\
Gemma-7B & 0.602 & 0.309 \\
Mistral-7B-Instruct-v0.3 & 0.777 & 0.369 \\
Llama 3.1 Instruct-8B & 0.832 & 0.374 \\
Qwen3-8B & 0.833 & 0.420 \\
Phi-4 & 0.912 & 0.491 \\
Qwen3-14B & 0.870 & 0.456 \\
Gemma 3 27B IT & 0.920 & 0.529 \\
Qwen3-32B & 0.892 & 0.481 \\
GPT-5.4 & 0.946 & 0.564 \\
Gemini 3.1 Flash Lite & 0.937 & \underline{0.680} \\
Claude Opus 4.6 & \underline{0.956} & 0.648 \\
\hline
\end{tabular}
\caption{Accuracy by question type on the GloBI portion of BAGEL ($n{=}3{,}117$). Open-weight models are ordered by increasing approximate parameter count. Best value in each column is \underline{underlined}.}
\label{tab:globi_breakdown}
\end{table*}

\subsection{Error analysis}
\label{subsec:error_analysis}

Headline accuracy alone does not explain \emph{why} models fail. We therefore pooled errors across the 14 models with complete predictions on all three domains (Table~\ref{tab:domain_accuracy}), including Gemini 3.1 Flash Lite, matching items by question text and gold answer when identifiers differ.

Table~\ref{tab:error_by_dimension} ranks all 11 BAGEL dimensions by pooled mean error rate (see also Appendix Figure~\ref{fig:dimension_error_rates}). GloBI Masked participant identification is the most difficult category (56.8\% mean error). Several Wikipedia dimensions, including morphology, behavior, communication, taxonomy, and cognition, cluster in the low-to-mid 30s, although Cognition contains only 25 items and its estimate should be interpreted cautiously. bioRxiv Result Interpretation is easier on average (16.4\% error), although a small number of items are incorrect for every model (Appendix Figure~\ref{fig:unanimous_failures}). This ranking aligns with the domain-level pattern in Table~\ref{tab:domain_accuracy}: ecological role assignment is harder than encyclopedic recall or short-form literature interpretation.

\begin{table*}[t]
\centering
\small
\setlength{\tabcolsep}{5pt}
\begin{tabular}{llrr}
\hline
\textbf{Domain} & \textbf{Dimension} & \textbf{Mean accuracy} & \textbf{Mean error rate} \\
\hline
GloBI & Masked participant identification & 0.4322 & 0.5678 \\
Wikipedia & Morphology & 0.6586 & 0.3414 \\
Wikipedia & Behavior & 0.6692 & 0.3308 \\
Wikipedia & Communication & 0.6779 & 0.3221 \\
Wikipedia & Taxonomy & 0.6953 & 0.3047 \\
Wikipedia & Cognition & 0.6964 & 0.3036 \\
Wikipedia & Diet & 0.7561 & 0.2439 \\
GloBI & Masked interaction type inference & 0.7726 & 0.2274 \\
Wikipedia & Geographic Distribution & 0.7822 & 0.2178 \\
Wikipedia & Habitat & 0.8138 & 0.1862 \\
bioRxiv & Result Interpretation & 0.8356 & 0.1644 \\
\hline
\end{tabular}
\caption{All BAGEL dimensions ranked by mean error rate across the $14$ complete models ($n{=}6{,}822$).}
\label{tab:error_by_dimension}
\end{table*}

Appendix Table~\ref{tab:globi_error_examples} provides illustrative mistakes on the hardest GloBI subtask. Incorrect answers are often biologically plausible rather than arbitrary. The examples include confusion among ecological roles (host, vector, or affected participant); near-miss substitutions among familiar congeners, such as selecting \textit{Aedes aegypti} instead of \textit{Aedes africanus}; and mistakes at the wrong taxonomic rank, such as species versus subspecies (\textit{Pteronotus parnellii} versus \textit{P.~parnellii fuscus}). These examples are consistent with models recognizing related taxa while failing to infer directional interaction structure reliably from a short stem.

At the item level, failures are unevenly distributed. Of the 6{,}822 questions in BAGEL, 85 are incorrect for every evaluated model; 76 of these are GloBI participant-identification items (Appendix Figure~\ref{fig:unanimous_failures}). At the other extreme, strong checkpoints contribute only a handful of model-exclusive errors (Appendix Table~\ref{tab:exclusive_failures}): Claude Opus~4.6, Gemma~3 27B IT, Llama~3.1-8B, and Qwen3-4B/32B have none by this count, whereas the smallest models accumulate hundreds of one-off errors. Validation removed ambiguous items that disproportionately depressed strong-model accuracy (Appendix Figure~\ref{fig:filter_gain}), but the unanimous-failure set suggests that some GloBI participant-identification items remain genuinely difficult even after audit.

\subsection{Post-release bioRxiv probe}
\label{subsec:fresh_biorxiv}

Training-data exposure complicates interpretation of literature-based benchmark scores. We therefore constructed a non-overlapping holdout from preprints posted on or after March~6, 2026, the release date of GPT-5.4. We queried the same animal-related bioRxiv categories, sampled 100 post-release preprints, concatenated each title, abstract, and body using the main pipeline, and generated one Result Interpretation question per document with the same GPT-4o-mini prompt (Appendix~\ref{app:biorxiv_bench_gen_prompt}). The holdout underwent the same GPT-5.5 filtering as the main benchmark. GPT-5.4 answered 96 of 100 retained holdout items correctly (96.0\%), close to 1{,}854/1{,}899 (97.6\%) on the main bioRxiv portion of BAGEL (Appendix Table~\ref{tab:fresh_biorxiv_gpt54}). This post-release probe suggests that strong bioRxiv performance is not explained solely by memorization of pre-release documents, while remaining a targeted check rather than a direct estimate of the model's training data.

\subsection{Takeaways for benchmark use}
\label{subsec:takeaways}

We find it more informative to interpret BAGEL as a collection of related tests than to rank models on a single aggregate score~\citep{hendrycks2021mmlu}. A model may appear capable on encyclopedic or literature-based items while still failing ecological interaction questions, and within GloBI the gap between interaction-type inference and participant identification is especially large. GPT-5.5 screening, calibrated against the expert-reviewed sample (Section~\ref{sec:human_validation}), removed items flagged as ambiguous or mislabeled before evaluation; residual issues may nevertheless remain in the released set.

\section{Conclusion}
\label{sec:conclusion}

We present BAGEL, a 6,822-question closed-book benchmark built from Wikipedia, GloBI, and bioRxiv, with GPT-5.5 screening calibrated on expert review. Across 14 models, performance depends strongly on source and question type: proprietary checkpoints lead the best open-weight systems overall, but even frontier models remain comparatively weak on GloBI participant identification. Pooled errors highlight ecological role confusion and taxonomic near-misses, while a small post-release bioRxiv probe provides preliminary evidence against purely verbatim recall.

\section*{Limitations}
\label{sec:limitations}

Several directions remain open. BAGEL is English-only and multiple-choice; extending it to additional languages, open-ended answers, or retrieval-augmented protocols would broaden what can be measured. Our main results use greedy decoding with a single seed, so studying alternative decoding settings is a natural next step for refining relative rankings. Finally, because the sources are public, closed-book scores may partly reflect familiarity with web text; the post-release bioRxiv probe is one check against memorization, and stronger leakage controls remain useful future work.

\section*{Ethical Considerations}
\label{sec:ethical_considerations}

BAGEL is intended as a research diagnostic of animal-related knowledge in language models, not as a substitute for domain experts, primary literature, or authoritative biodiversity databases. Benchmark accuracy can be overgeneralized: a model that performs well in aggregate may still make systematic ecological-role or taxonomic errors. If such outputs were used without verification in ecological research, conservation decisions, education, or public communication, they could propagate false biological claims. We therefore report results by source and question type, document known failure modes, and recommend checking consequential claims against authoritative sources. The English-only, multiple-choice design and reliance on three public sources also underrepresent other languages, knowledge traditions, taxa, and forms of biological expertise.

The internal construction files include source text needed to generate and audit questions. To avoid republishing those documents, the public release contains the benchmark questions, answers, source identifiers, and associated metadata, while users obtain source documents from their original hosts. Before release, all retained public fields were screened automatically for indicators of personally identifying and objectionable content, and every flagged record was manually reviewed; no such content was retained. This screening reduces but cannot guarantee the absence of every undesirable string. Construction, model-based validation, and evaluation also consumed API and GPU computation. Appendix~\ref{app:artifact_reproducibility} reports the recoverable computational record and implementation details so that this cost is visible.

\section*{Acknowledgments}
Jiacheng Shen is supported by the Earth Species Project. Computing resources were supported by the NYU Torch Cluster and the NYU Shanghai Frontier AI Center HPC cluster.

\bibliography{bagel}

\appendix
\section{Related Benchmarks}
\begin{table*}[t]
\centering
\scriptsize
\setlength{\tabcolsep}{4pt}
\renewcommand{\arraystretch}{1.12}
\begin{tabular}{p{2.2cm} p{2.45cm} p{2.5cm} p{4.9cm}}
\toprule
\textbf{Benchmark / line} & \textbf{Primary setting} & \textbf{Relation to animals / biodiversity} & \textbf{Key contrast with BAGEL} \\
\midrule
\textbf{BAGEL} &
Text-only, closed-book 4-option MC from three public sources &
Directly centered on animal and natural history knowledge &
\textbf{Reference row:} Wikipedia species facts, GloBI interaction reasoning, and bioRxiv literature questions under one closed-book protocol. \\
\midrule
MMLU~\citep{hendrycks2021mmlu} &
Broad text MC across 57 subjects &
Only incidental biology / life-science coverage &
BAGEL is animal-centered and reports scores by source rather than as one academic average. \\
ScienceQA~\citep{lu2022learn} &
Multimodal school-science MC &
Includes natural-science curriculum content &
BAGEL targets higher-level natural-history content and uses text-only closed-book questions built from biodiversity sources. \\
BEANS; NatureLM-audio~\citep{hagiwara2022beans,robinson2024naturelmaudio} &
Bioacoustic audio tasks; audio--language modeling &
Directly about animal sounds &
BEANS evaluates audio understanding; BAGEL evaluates text-only closed-book knowledge about animals. \\
Ecological knowledge eval~\citep{dorm2025ecologicalknowledge} &
Textual evaluation of ecological LLM competence (mixed task types) &
Shared theme: biodiversity-related scientific knowledge in language models &
BAGEL focuses on animal natural history in one closed-book format across three text sources; Dorm et al.\ use a broader mix of ecological tasks. \\
EnviroExam; ELLE~\citep{huang2024enviroexam,guo2025elle} &
Environmental-domain exams / QA &
Broad environment and eco-environment coverage &
BAGEL is narrower but deeper on animal and natural-history knowledge, with explicit species, interaction, and literature structure. \\
PubMedQA; BLURB~\citep{jin2019pubmedqa,gu2021domain} &
Biomedical QA / biomedical NLP &
Biological, but centered on human health and biomedicine &
BAGEL shifts the focus from biomedical language understanding to organismal diversity, ecology, and natural-history reasoning. \\
\bottomrule
\end{tabular}
\caption{Representative neighboring benchmarks and how BAGEL differs. Rows are illustrative, not exhaustive.}
\label{tab:related_benchmarks}
\end{table*}
\section{Dataset composition and difficulty strata}
\label{app:composition}
\begin{table*}[t]
\centering
\footnotesize
\setlength{\tabcolsep}{4.5pt}
\renewcommand{\arraystretch}{0.95}
\begin{tabular}{llrr}
\hline
\textbf{Domain} & \textbf{Dimension} & \textbf{\# Questions} & \textbf{Share in Domain} \\
\hline
\multicolumn{4}{c}{\textbf{bioRxiv} (\# = 1,899)} \\
\hline
bioRxiv & Result Interpretation & 1,899 & 100.0\% \\
\hline
\multicolumn{4}{c}{\textbf{GloBI} (\# = 3,117)} \\
\hline
GloBI & Masked interaction type inference & 1,888 & 60.6\% \\
GloBI & Masked participant identification & 1,229 & 39.4\% \\
\hline
\multicolumn{4}{c}{\textbf{Wikipedia} (\# = 1,806)} \\
\hline
Wikipedia & Behavior & 331 & 18.3\% \\
Wikipedia & Diet & 278 & 15.4\% \\
Wikipedia & Geographic Distribution & 264 & 14.6\% \\
Wikipedia & Taxonomy & 251 & 13.9\% \\
Wikipedia & Communication & 219 & 12.1\% \\
Wikipedia & Habitat & 219 & 12.1\% \\
Wikipedia & Morphology & 219 & 12.1\% \\
Wikipedia & Cognition & 25 & 1.4\% \\
\hline
\end{tabular}
\caption{Composition of BAGEL across source domains and source-specific dimensions ($n{=}6{,}822$). Dimension shares within each domain follow the generation proportions; domain totals reflect validated counts. Mean question length is 180.5 characters and mean option length is 35.7 characters.}
\label{tab:bagel_full_stats}
\end{table*}
\begin{table*}[t]
\centering
\footnotesize
\setlength{\tabcolsep}{5pt}
\renewcommand{\arraystretch}{0.95}
\begin{tabular}{lrrrr}
\hline
\textbf{Domain} & \textbf{\# Items} & \textbf{Easy} & \textbf{Medium} & \textbf{Hard} \\
\hline
Wikipedia & 1,806 & 1,567 (86.8\%) & 191 (10.6\%) & 48 (2.7\%) \\
GloBI & 3,117 & 2,210 (70.9\%) & 618 (19.8\%) & 289 (9.3\%) \\
bioRxiv & 1,899 & 1,762 (92.8\%) & 114 (6.0\%) & 23 (1.2\%) \\
\hline
\textbf{All} & \textbf{6,822} & \textbf{5,539 (81.2\%)} & \textbf{923 (13.5\%)} & \textbf{360 (5.3\%)} \\
\hline
\end{tabular}
\caption{Reference difficulty strata in BAGEL ($n{=}6{,}822$), induced by GPT-5.4, Claude Opus~4.6, and Gemini 3.1 Flash Lite: \textbf{easy} if all three models answer correctly, \textbf{medium} if exactly one or two are correct, and \textbf{hard} if none is correct.}
\label{tab:bagel_difficulty_three_frontier}
\end{table*}

\section{Artifact documentation and reproducibility}
\label{app:artifact_reproducibility}

\paragraph{Coverage and intended use.}
The released artifact is available at \url{https://huggingface.co/datasets/EarthSpeciesProject/BAGEL}. It is an English-language evaluation benchmark with 6{,}822 four-option questions. It contains no training or development split: all three source partitions are evaluation sets. Source- and dimension-level counts are reported in Table~\ref{tab:bagel_full_stats}. The intended use is research evaluation and error analysis of animal-related language-model knowledge. It is not designed for training models, making claims about biological truth without source verification, or automating consequential ecological or conservation decisions.

\paragraph{Source access, attribution, and release terms.}
Wikipedia text is available under CC BY-SA 4.0 and, where applicable, the GNU Free Documentation License;\footnote{\url{https://en.wikipedia.org/wiki/Wikipedia:Copyrights}} GloBI states that its integrated data are available under CC BY 4.0 unless otherwise noted and asks reusers to preserve attribution to the original contributing datasets;\footnote{\url{https://www.globalbioticinteractions.org/data}} and bioRxiv explicitly permits text and data mining while leaving redistribution and derivative-use terms to each article's author-selected license.\footnote{\url{https://www.biorxiv.org/tdm}} We used these sources for research-oriented question construction. Internal construction records contain source text, but the public release excludes Wikipedia extracts, GloBI source tables, and bioRxiv full text. Public records contain the generated question, four options, gold answer, source domain, question dimension, and minimal provenance needed to trace construction. Record-level GloBI dataset and reference provenance is preserved where supplied. Users seeking the underlying source text must obtain it from the original host under the applicable terms. Code is released under Apache License 2.0. The questions are released under CC BY-SA 4.0, and the dataset page records source attribution. bioRxiv redistribution remains subject to each article's license.

\paragraph{Privacy and content screening.}
BAGEL targets animals and ecological interactions and does not intentionally collect data about people. Although internal construction records retain source text, the public release will exclude those fields and author contact information. We ran automated checks over all retained public textual fields for indicators of personally identifying information and objectionable content, followed by manual review of every flagged record. No flagged personal identifier or offensive item was retained. This check was separate from the 150-item scientific-quality audit in Section~\ref{sec:human_validation}.

\paragraph{Models and implementation.}
For reproducibility, the open-weight checkpoints were fixed to the following Hugging Face revisions: SmolLM2-360M-Instruct (\texttt{a10cc1512}), Qwen3-0.6B (\texttt{c1899de2}), Qwen3-4B (\texttt{1cfa9a72}), Qwen3-8B (\texttt{b968826d}), Qwen3-14B (\texttt{40c06982}), Gemma-7B (\texttt{ff6768d9}), Gemma-3-27B-IT (\texttt{005ad340}), Llama-3.1-8B-Instruct (\texttt{0e9e39f2}), Mistral-7B-Instruct-v0.3 (\texttt{c170c708}), and Phi-4 (\texttt{932b33c0}). The Qwen3-32B run used the \texttt{Qwen/Qwen3-32B} repository; its exact cached revision was not retained unambiguously with the prediction output. Open-weight inference used PyTorch~2.11.0, Transformers~5.3.0, Accelerate~1.13.0, Datasets~4.8.4, and bfloat16 weights, with one model per NVIDIA A100 80\,GB GPU. The evaluation truncated tokenized inputs at 1{,}000 tokens, generated at most 16 new tokens, disabled sampling and Qwen thinking mode, and used the prompt in Appendix~\ref{app:evaluation_prompt}. API inference used OpenAI~2.29.0 together with the provider interfaces for the exact model identifiers reported in Section~\ref{sec:experimental_setup}.

\paragraph{Computational budget.}
Each open-weight checkpoint was evaluated once on all 6{,}822 items; no training or hyperparameter search was performed. API models were likewise evaluated once per item. Recoverable Slurm accounting records, including unsuccessful and development jobs, account for at least 9.2 NVIDIA A100 80\,GB GPU-hours; this is a lower bound because complete accounting for some early development runs was not retained. Generation and GPT-5.5 filtering used hosted APIs rather than local GPUs. Proprietary-model parameter counts and provider-side hardware are not publicly disclosed. We report this lower bound rather than presenting an incomplete record as an exact total.

\section{Author audit protocol}
\label{app:human_audit_protocol}

The sole human auditor was an author of the submission. For each sampled item, the auditor was shown the source context, question, four answer options, and proposed gold answer. The annotation was binary and multilabel, with the following codebook:
\begin{enumerate}[leftmargin=*]
    \item \emph{good}: the item is valid and answerable as written;
    \item \emph{identical or near-identical options}: at least two options are semantic duplicates;
    \item \emph{multiple plausible answers}: more than one option is plausibly correct given the context or relevant common knowledge;
    \item \emph{incorrect gold answer}: the provided gold answer appears wrong;
    \item \emph{unsupported by context}: the answer cannot be justified by the supplied context or evidence; and
    \item \emph{format issue}: the option set, prompt, or required fields are malformed.
\end{enumerate}
Labels were judged independently, and \emph{good} was used only when no substantial issue was present. For the released benchmark, the operational remove-or-keep rule used the two labels calibrated in Section~\ref{sec:human_validation}: an item was removed if either \emph{multiple plausible answers} or \emph{incorrect gold answer} was true. The task involved quality control of public, machine-generated scientific questions, collected no information about the auditor, and used no external participants.

\section{Appendix}

\subsection{Prompts}
We used the following prompts in this project.
\subsubsection{Benchmark Generation Prompt}
\label{wiki_bench_gen_prompt}
\begin{promptbox}
    {Wikipedia domain benchmark generation prompt}

System: You are a scientific question generator.
Given a Wikipedia article about a single species, generate up to eight multiple-choice questions (no free-form).
Only use information explicitly mentioned in the text.

Each question must:
- Fall under one of these dimensions: Taxonomy, Behavior, Communication, Morphology, Habitat, Cognition, Geographic Distribution, Diet.
- For Behavior, consider social behavior (e.g., solitary vs. social, group size, social structure) when present.
- Have exactly four options and exactly one correct answer.
- Be factual, unambiguous, and grounded in the given Wikipedia text (no outside knowledge).
- Skip dimensions not covered in the text.

If the text mentions vocalization and communication, include at least one question about Communication.
If the text mentions where the species is found (e.g., native range, regions, countries), include at least one question about Geographic Distribution.
If the text mentions feeding or diet, include at least one question about Diet.\n\n
User:Species metadata:
ScientificName: {ScientificName}
CommonNames: {CommonNames}
Taxonomy: Kingdom={Kingdom}; Phylum={Phylum}; Class={Class}; Order={Order}; Family={Family}; Genus={Genus}
WikipediaTitle: {WikipediaTitle}

WikipediaText:
\"\"\"{WikipediaText}\"\"\"

Instructions:
Generate up to EIGHT multiple-choice questions across relevant dimensions.
Return output as a JSON list under this schema:

{{
  "questions": [
    {{
      "dimension": "<one of Taxonomy|Behavior|Communication|Morphology|Habitat|Cognition|Geographic Distribution|Diet>",
      "question": "<question text>",
      "options": ["<opt1>", "<opt2>", "<opt3>", "<opt4>"],
      "answer": "<correct option>"
    }}
  ]
}}\n\n
Assistant:\n\n
\end{promptbox}

\label{app:globi_bench_gen_prompt}
\begin{promptbox}{GloBI domain benchmark generation prompt}
System: You are creating benchmark-quality self-contained multiple-choice questions. Return only one valid JSON object
User: You are generating benchmark-quality multiple-choice questions for the ecological interaction domain of an animal expertise benchmark.

You will be given a short source context derived from a GloBI ecological interaction record.

Your task is to generate exactly ONE self-contained multiple-choice question.

The benchmark is CLOSED-BOOK:
- The final test-taker will NOT see the source context.
- The question must stand on its own.
- The question must NOT refer to "the passage", "the text", "the record", "the source", or any hidden context.
- Use ONLY information supported by the provided source context. Do NOT use outside knowledge.

Your goal is to produce a question that tests ecological reasoning, not keyword matching.

Select exactly ONE dimension that is most suitable for the source context:
- Masked interaction type inference
- Masked participant identification

Dimension guidance:
- Masked participant identification: hide one participant and ask which taxon best fills the missing ecological slot.
- Masked interaction type inference: hide the relation label and ask which interaction type best fits the interaction pattern.

High-value question criteria:
- The item must require at least one inference step.
- The correct answer should NOT be recoverable from a single cue word alone.
- The species names and ecological clues should matter to solving the question.
- The question should test ecological understanding, not simple paraphrase or label recognition.

Anti-leakage rules:
- Do NOT use verbs or phrases that directly reveal the correct interaction type, such as:
  "consumes", "eats", "predates on", "parasitizes", "pollinates", "infects", "hosts", "mutualist", "commensal", "parasite", "predator", "prey".
- Do NOT use wording that is an obvious synonym or near-restatement of the correct answer.
- Do NOT write a question that can be answered correctly even if the taxa are replaced with "species X" and "species Y".
- Do NOT make the correct answer obvious from one lexical clue in the stem.
- If the source context only supports a trivial label-retrieval question, reject it.
For masked interaction type inference, do not ask for the interaction label directly unless the stem requires integrating multiple clues.
Prefer questions where the solver must infer the interaction from ecological roles, asymmetry, or biological consequences, not from a single action verb.
Requirements for the question:
- Include only the clues necessary to support a non-trivial inference.
- Prefer ecological reasoning over direct restatement.
- Do not simply restate the source context verbatim unless necessary.
- Do not overgeneralize from a single documented interaction.
- For masked-inference questions, the missing field must be reasonably inferable from the remaining clues.
- Avoid questions where multiple options could plausibly be correct.
- Use clear scientific language.
- The question should read like an ecological reasoning problem, not like a hidden database entry.

Requirements for options:
- Exactly four options labeled A, B, C, and D.
- Exactly one correct answer.
- Distractors must be plausible.
- Distractors should reflect realistic ecological confusions, such as:
  - reversed interaction direction
  - wrong ecological role
  - wrong interaction type
  - plausible but incorrect taxon
  - unsupported stronger claim

Reject low-value questions such as:
- direct label retrieval from a cue word
- simply copying a species name with no reasoning value
- masked questions where the answer is not actually inferable from the clues
- items where the stem already contains the answer in nearly explicit form
- items that test generic ecology vocabulary rather than animal expertise

Before deciding the item is usable, internally check:
1. Can the answer be solved from a single cue word or phrase?
2. Would the item still work the same if the species names were replaced with generic placeholders?
3. Is the correct option mostly a paraphrase of the stem?
4. Does the item require ecological reasoning rather than lexical matching?

If the answer to any of 1-3 is yes, strongly prefer rejection.

Avoid wording such as:
- "documented"
- "recorded"
- "reported"
- "observed"
- "in the record"
- "according to the source"
- "the passage states"

Return valid JSON in one of these two formats.

USABLE:
{
  "usable": "true",
  "dimension": "<one of: Masked participant identification | Masked interaction type inference>",
  "question": "<self-contained question>",
  "options": {
    "A": "<option A>",
    "B": "<option B>",
    "C": "<option C>",
    "D": "<option D>"
  },
  "correct_answer": "<A|B|C|D>",
  "supporting_evidence": [
    "<short evidence span 1>",
    "<short evidence span 2>"
  ],
  "rationale": "<brief explanation grounded in the source context>"
}

REJECTION:
{
  "usable": "false",
  "rejection_reason": "<one of: too_sparse | answer_not_self_contained | low_reasoning_value | masked_field_not_inferable | lexical_leakage>"
}

SOURCE CONTEXT:
{Interaction record}\n\n Assistant:\n\n
\end{promptbox}
\label{app:biorxiv_bench_gen_prompt}
\begin{promptbox}{BioRxiv domain benchmark generation prompt}
System: You are creating benchmark-quality self-contained multiple-choice questions. Return only one valid JSON object.
User: You are creating benchmark-quality multiple-choice questions from a bioRxiv paper.

The final benchmark model will see ONLY the question and answer options. Therefore, the question itself must include all necessary context for answering.

Read the paper and generate exactly ONE self-contained 4-choice multiple-choice question.

Use ONLY information from the paper. Do NOT use outside knowledge.
Only generate the following type of dimension questions:
- Result Interpretation --- interpreting experimental results

Requirements:
- The question must be fully self-contained.
- Include the minimum necessary scientific context inside the question stem so that the question is answerable without access to the paper.
- The question stem may be longer than usual, but it must remain concise and focused.
- The question stem must include setup and evidence when needed, but must NOT directly restate the correct answer or final conclusion in answer-like wording.
- Do NOT create a question that can be answered only by having seen the paper.
- The question must test scientific reasoning, not trivial word matching.
- Distractors must be scientifically plausible and belong to the same conceptual category as the correct answer.
- Provide exactly four options labeled A, B, C, and D.
- Exactly one option must be correct.
- Do not mention the dimension in the question text.
- Do not include unsupported information.

Question-writing guidance:
- Prefer a stem that includes:
  1. the study system or biological setting,
  2. the relevant manipulation, comparison, or observation,
  3. the key evidence or result pattern,
  4. the reasoning task.
- Avoid copying long sentences directly from the paper.
- Lightly rewrite the source information into a natural, self-contained scientific scenario.
- Do not make the stem so compressed that the correct answer becomes obvious from wording alone.

Before finalizing, check:
- Can a model answer the question from the stem alone?
- Does the stem avoid directly revealing the correct answer?
- Are the options plausible and same-category?
- Is the question testing reasoning rather than recall of the paper?

Return valid JSON in this schema:

{
  "dimension": "Result Interpretation ",
  "question": "<self-contained question stem>",
  "options": {
    "A": "<option A>",
    "B": "<option B>",
    "C": "<option C>",
    "D": "<option D>"
  },
  "correct_answer": "<A|B|C|D>",
  "supporting_evidence": [
    "<short copied or lightly trimmed evidence span 1>",
    "<short copied or lightly trimmed evidence span 2>"
  ],
  "reason": "<brief explanation of why the correct answer is supported by the paper>"
}

PAPER:
{paper}\n\n
Assistant:\n\n
\end{promptbox}

\subsubsection{Evaluation Prompt}
\label{app:evaluation_prompt}

\begin{promptbox}{Evaluation prompt}
You are answering a multiple-choice closed-book benchmark question for testing animal expertise.
Choose exactly one answer.\n
Output exactly one capital letter: A, B, C, or D.\n
Do not output any explanation, words, punctuation, or extra text.\n\n
Question: {question}\n
Options:
A. {option_a}
B. {option_b}
C. {option_c}
D. {option_d}
Answer:
\end{promptbox}

\input{tables/benchmark_examples_main}
\input{tables/benchmark_examples_appendix}
\section{Wikipedia accuracy by dimension}
\label{app:wiki_breakdown}

Table~\ref{tab:globi_breakdown} in Section~\ref{subsec:question_type_accuracy} reports GloBI question-type accuracy in the main text. Table~\ref{tab:wiki_breakdown} gives the corresponding Wikipedia breakdown ($n{=}1{,}806$).
\begin{table*}[t]
\centering
\scriptsize
\setlength{\tabcolsep}{4pt}
\begin{tabular}{lcccccccc}
\hline
\textbf{Model} & \textbf{Behavior} & \textbf{Cognition} & \textbf{Communication} & \textbf{Diet} & \textbf{Geographic Distribution} & \textbf{Habitat} & \textbf{Morphology} & \textbf{Taxonomy} \\
\hline
SmolLM2-360M-Instruct & 0.238 & 0.250 & 0.223 & 0.232 & 0.242 & 0.218 & 0.289 & 0.226 \\
Qwen3-0.6B & 0.232 & 0.167 & 0.232 & 0.225 & 0.239 & 0.253 & 0.289 & 0.270 \\
Qwen3-4B & 0.653 & 0.667 & 0.664 & 0.732 & 0.773 & 0.817 & 0.606 & 0.597 \\
Gemma-7B & 0.539 & 0.667 & 0.559 & 0.704 & 0.761 & 0.852 & 0.550 & 0.690 \\
Mistral-7B-Instruct-v0.3 & 0.641 & 0.667 & 0.700 & 0.754 & 0.792 & 0.838 & 0.541 & 0.621 \\
Llama 3.1 Instruct-8B & 0.715 & 0.708 & 0.714 & 0.782 & 0.811 & 0.860 & 0.674 & 0.706 \\
Qwen3-8B & 0.740 & 0.708 & 0.741 & 0.839 & 0.845 & 0.900 & 0.697 & 0.677 \\
Phi-4 & 0.768 & 0.875 & 0.768 & 0.861 & 0.856 & 0.934 & 0.702 & 0.774 \\
Qwen3-14B & 0.752 & 0.833 & 0.736 & 0.854 & 0.886 & 0.930 & 0.739 & 0.742 \\
Gemma 3 27B IT & 0.728 & 0.833 & 0.782 & 0.907 & 0.879 & 0.926 & 0.720 & 0.782 \\
Qwen3-32B & 0.762 & 0.667 & 0.777 & 0.879 & 0.890 & 0.943 & 0.729 & 0.794 \\
GPT-5.4 & 0.842 & 0.875 & 0.850 & 0.932 & 0.989 & 0.969 & 0.872 & 0.940 \\
Gemini 3.1 Flash Lite & 0.858 & 0.875 & 0.850 & 0.936 & 0.996 & 0.969 & 0.917 & 0.944 \\
Claude Opus 4.6 & \underline{0.901} & \underline{0.958} & \underline{0.895} & \underline{0.950} & \underline{0.992} & \underline{0.983} & \underline{0.894} & \underline{0.972} \\
\hline
\end{tabular}
\caption{Accuracy by question type on the Wikipedia portion of BAGEL. Open-weight models are ordered by increasing approximate parameter count. Best value in each column is \underline{underlined}.}
\label{tab:wiki_breakdown}
\end{table*}

\section{Supplementary error-analysis details}
\label{app:error_analysis}

\begin{figure*}[t]
\centering
\includegraphics[width=0.88\linewidth]{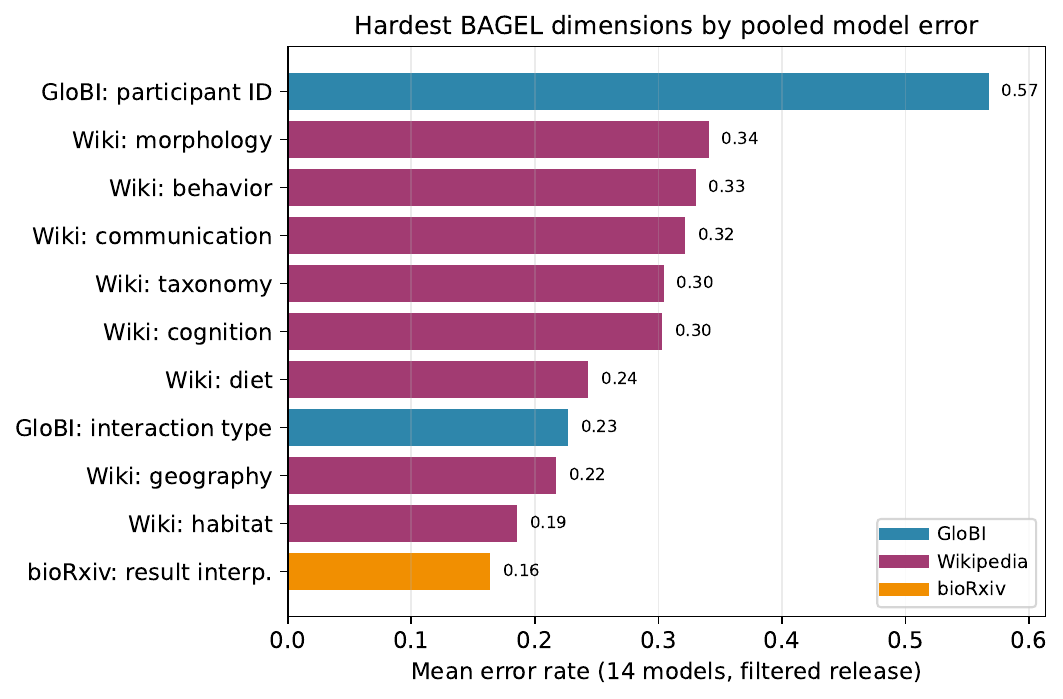}
\caption{Mean error rate by BAGEL dimension, averaged across the $14$ complete models ($n{=}6{,}822$). Colors indicate source domain.}
\label{fig:dimension_error_rates}
\end{figure*}
\begin{table*}[t]
\centering
\footnotesize
\setlength{\tabcolsep}{4pt}
\begin{tabular}{p{2.0cm}p{5.0cm}p{2.3cm}p{2.3cm}}
\hline
\textbf{ID} & \textbf{Question (short)} & \textbf{Gold} & \textbf{Common wrong answer} \\
\hline
\texttt{globi:359:0} & Likely vector for Zika virus in the observed interaction & \textit{Aedes africanus} & \textit{Aedes aegypti} \\
\texttt{globi:303:0} & Likely host where \textit{Strebla altmani} is involved & \textit{Pteronotus parnellii fuscus} & \textit{Pteronotus parnellii} \\
\texttt{globi:14:0} & Taxon impacted when \textit{Parvioris mortoni} negatively affects another organism & \textit{Archaster typicus} & \textit{Echinometra lucunter} \\
\texttt{globi:1887:0} & Likely host for endoparasite \textit{Syngamus trachea} & \textit{Pica pica} & \textit{Passer domesticus} \\
\texttt{globi:1206:0} & Likely host involving \textit{Crenosoma vulpis} & \textit{Cepaea nemoralis} & \textit{Canis vulpes} \\
\hline
\end{tabular}
\caption{Representative shared errors on GloBI \textit{Masked participant identification}. ``Common wrong answer'' is the most frequent incorrect option among the $14$ evaluated models.}
\label{tab:globi_error_examples}
\end{table*}
\begin{figure}[t]
\centering
\includegraphics[width=0.72\linewidth]{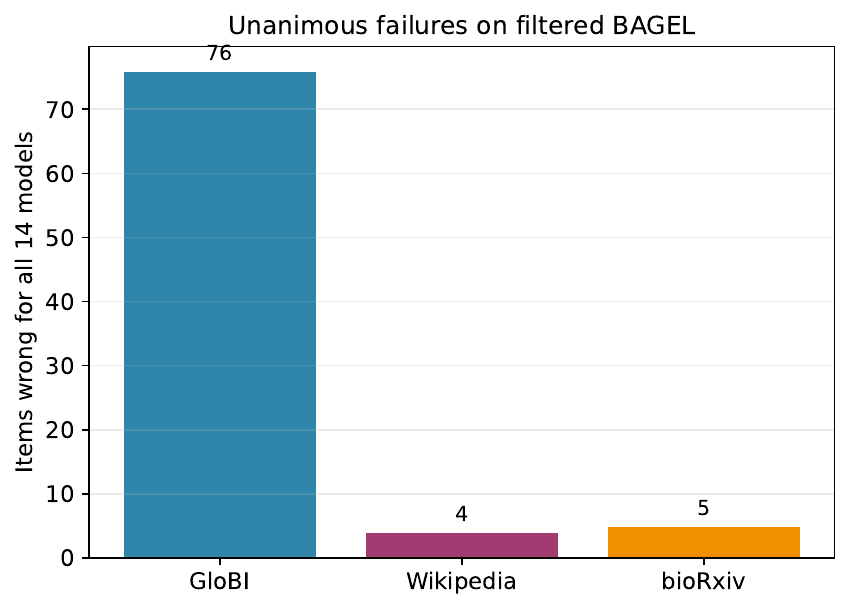}
\caption{Items answered incorrectly by all $14$ complete models on BAGEL, by source domain.}
\label{fig:unanimous_failures}
\end{figure}

\begin{table}[t]
\centering
\small
\setlength{\tabcolsep}{4pt}
\begin{tabular}{lr}
\hline
\textbf{Model} & \textbf{Exclusive failures} \\
\hline
SmolLM2-360M-Instruct & 902 \\
Qwen3-0.6B & 662 \\
Mistral-7B-Instruct-v0.3 & 3 \\
Qwen3-8B & 2 \\
GPT-5.4 & 2 \\
Gemini 3.1 Flash Lite & 1 \\
Phi-4 & 1 \\
Qwen3-14B & 1 \\
All other models & 0 \\
\hline
\end{tabular}
\caption{Model-exclusive failures: items where exactly one of the $14$ complete models is wrong.}
\label{tab:exclusive_failures}
\end{table}

\begin{figure}[t]
\centering
\includegraphics[width=0.88\linewidth]{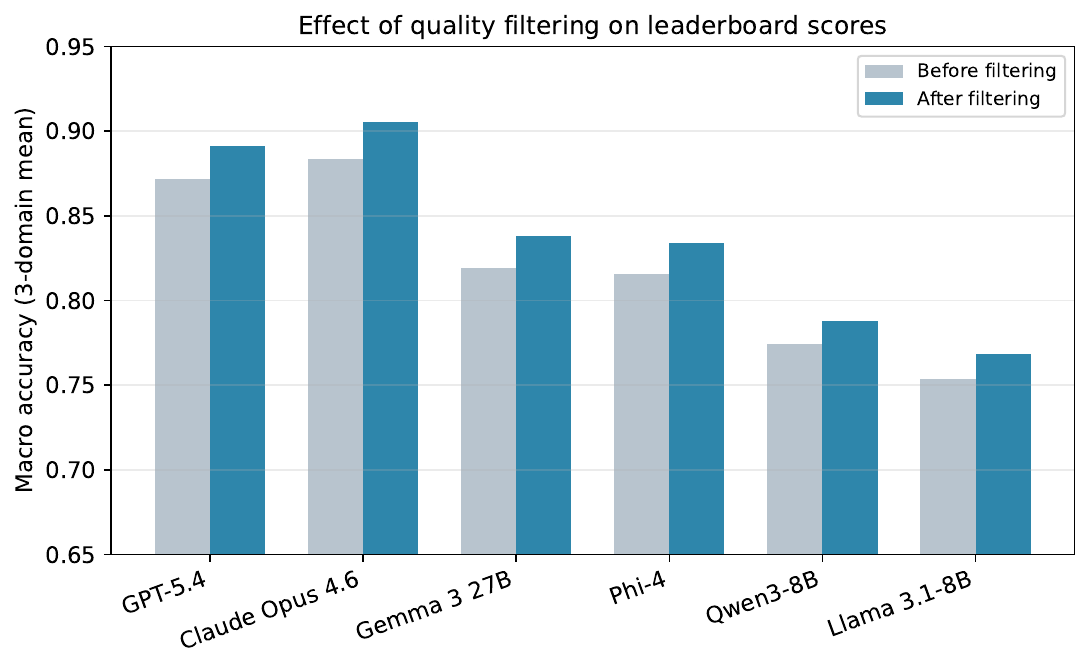}
\caption{Three-domain macro accuracy before versus after GPT-5.5 screening calibrated on expert review, for representative models.}
\label{fig:filter_gain}
\end{figure}

The main error-analysis results appear in Section~\ref{subsec:error_analysis}. Figure~\ref{fig:filter_gain} shows that removing validation-flagged questions increases three-domain macro accuracy for strong models because ambiguous items were disproportionately difficult; frontier rankings remain qualitatively similar, but the validated dataset provides a tighter evaluation.

\section{Post-release bioRxiv subset}
\label{app:fresh_biorxiv}

The post-release bioRxiv holdout experiment is summarized in Section~\ref{subsec:fresh_biorxiv}. We queried the bioRxiv API for animal-related categories (Animal Behavior and Cognition, Ecology, Evolutionary Biology, and Zoology), sampled 100 non-overlapping preprints posted on or after March~6, 2026, and processed each title, abstract, and body using the main pipeline. We generated one Result Interpretation question per document with the same GPT-4o-mini prompt (Appendix~\ref{app:biorxiv_bench_gen_prompt}) and applied the same GPT-5.5 filter used for the main benchmark.

\begin{table}[t]
\centering
\footnotesize
\setlength{\tabcolsep}{4pt}
\begin{tabular}{lrrr}
\hline
\textbf{Subset} & \textbf{$n$} & \textbf{Correct} & \textbf{Accuracy} \\
\hline
Main bioRxiv (BAGEL) & 1,899 & 1,854 & 97.6\% \\
Post-release holdout & 100 & 96 & 96.0\% \\
\hline
\end{tabular}
\caption{GPT-5.4 accuracy on the main bioRxiv portion of BAGEL and a post-release holdout of 100 animal-related preprints posted on or after March~6, 2026.}
\label{tab:fresh_biorxiv_gpt54}
\end{table}

\section{MMLU-Pro versus BAGEL}
\label{app:mmlu_vs_bagel}
\begin{figure}[t]
\centering
\includegraphics[width=0.88\linewidth]{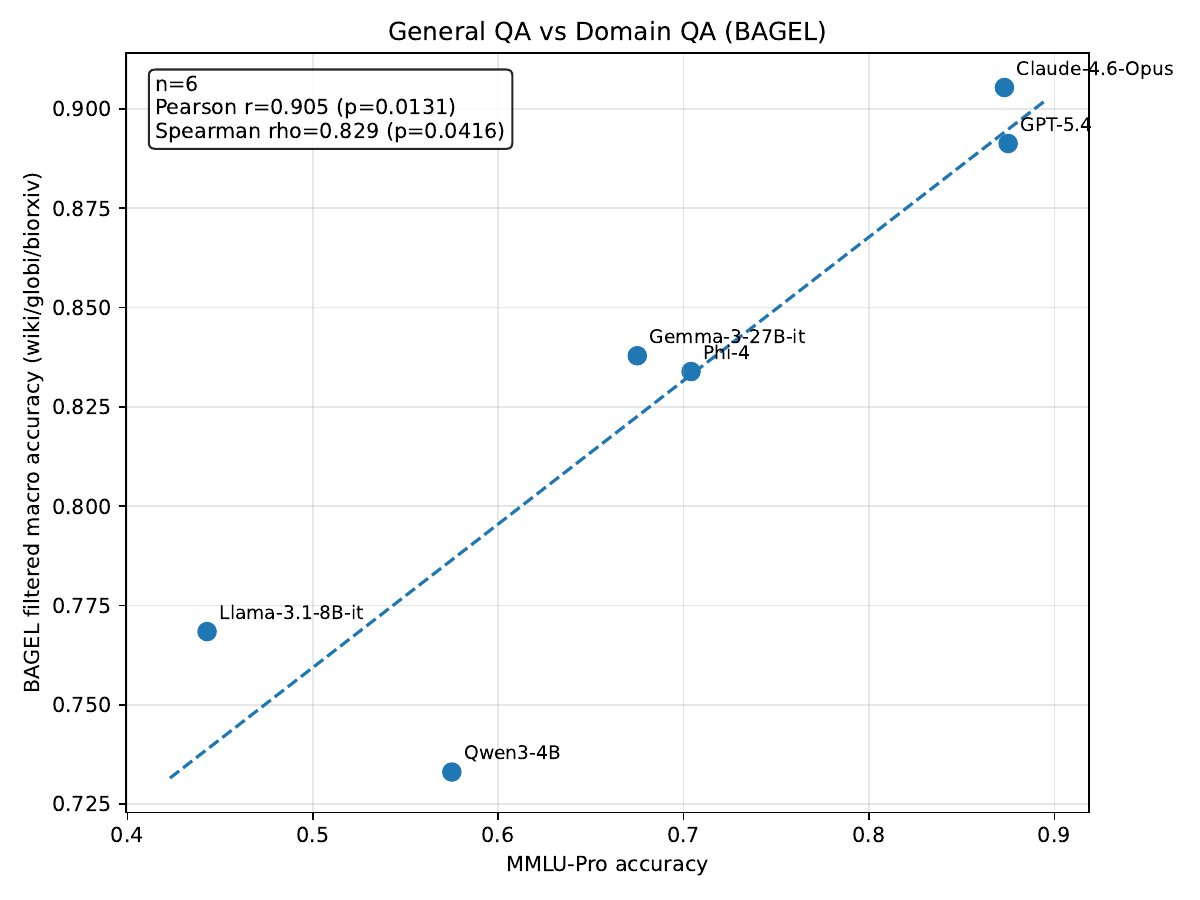}
\caption{MMLU-Pro versus BAGEL macro accuracy for six models with both reported scores. The dashed line is a linear fit.}
\label{fig:mmlu_vs_bagel}
\end{figure}

For six models with published MMLU-Pro scores, we plot BAGEL macro accuracy against MMLU-Pro (Figure~\ref{fig:mmlu_vs_bagel}). The two metrics correlate strongly (Pearson $r{=}0.91$, $p{=}0.013$; Spearman $\rho{=}0.83$, $p{=}0.042$), but they do not rank models identically: Llama~3.1-8B scores lower on MMLU-Pro than Qwen3-4B (0.44 vs.\ 0.58) yet higher on BAGEL (0.77 vs.\ 0.73). General academic knowledge and animal-domain knowledge overlap, but they are not the same thing.

\section{Qwen3 family performance on BAGEL}
\begin{figure*}[t]
\centering
\includegraphics[width=0.92\linewidth]{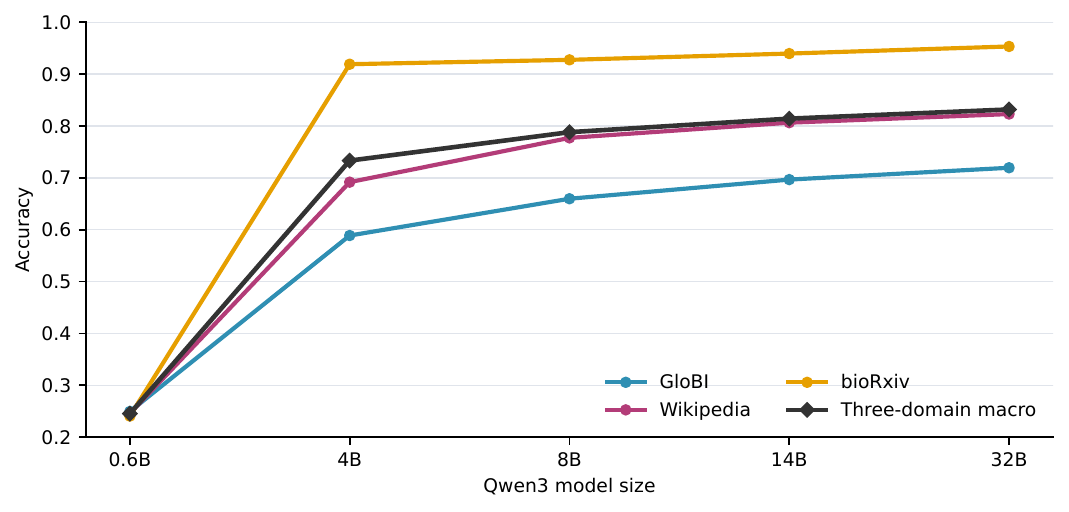}
\caption{Qwen3 instruct models (seed~0, greedy) on BAGEL: accuracy by domain and three-domain macro average across model sizes.}
\label{fig:qwen3_family_curves}
\end{figure*}

Figure~\ref{fig:qwen3_family_curves} plots Qwen3 instruct models from 0.6B to 32B on BAGEL.
\section{Answer-position bias before shuffling}
\label{app:answer_position_bias}

Before we shuffled answer options, the correct letter was unevenly distributed within each domain (Table~\ref{tab:appendix-answer-position-dataset}). Small models could exploit that skew---for example, by always answering ``B'' when most gold labels were B. After permutation, each position holds about a quarter of correct answers (Table~\ref{tab:appendix-answer-position-after-shuffle}).

Table~\ref{tab:appendix-answer-position-models} shows whether model outputs mirrored the same bias. Qwen3-0.6B almost always emits A; SmolLM2-360M-Instruct favors B. Larger models spread their answers more evenly but still show positional habits, especially on Wikipedia. Some high scores on the unshuffled set therefore reflected letter preference as much as content knowledge; shuffling removes that shortcut.

\begin{table*}[t]
\centering
\small
\setlength{\tabcolsep}{8pt}
\begin{tabular}{lrrrr}
\hline
\textbf{Domain} & \textbf{A} & \textbf{B} & \textbf{C} & \textbf{D} \\
\hline
bioRxiv & 659 (30.2\%) & 1,143 (52.4\%) & 325 (14.9\%) & 56 (2.6\%) \\
GloBI & 2,135 (61.0\%) & 658 (18.8\%) & 633 (18.1\%) & 74 (2.1\%) \\
Wikipedia & 199 (10.3\%) & 1,244 (64.6\%) & 445 (23.1\%) & 39 (2.0\%) \\
\hline
\end{tabular}
\caption{Correct-answer position distribution in the generated pool before option shuffling and GPT-5.5 screening.}
\label{tab:appendix-answer-position-dataset}
\end{table*}

\begin{table*}[t]
\centering
\small
\setlength{\tabcolsep}{8pt}
\begin{tabular}{lrrrr}
\hline
\textbf{Domain} & \textbf{A} & \textbf{B} & \textbf{C} & \textbf{D} \\
\hline
bioRxiv & 519 (23.8\%) & 556 (25.5\%) & 551 (25.2\%) & 557 (25.5\%) \\
GloBI & 883 (25.2\%) & 877 (25.1\%) & 869 (24.8\%) & 871 (24.9\%) \\
Wikipedia & 487 (25.3\%) & 479 (24.9\%) & 490 (25.4\%) & 471 (24.4\%) \\
\hline
All domains & 1,889 (24.8\%) & 1,912 (25.1\%) & 1,910 (25.1\%) & 1,899 (25.0\%) \\
\hline
\end{tabular}
\caption{Correct-answer position distribution in the generated pool \emph{after} option shuffling ($n{=}7{,}610$). GPT-5.5 screening subsequently retained 6{,}822 items.}
\label{tab:appendix-answer-position-after-shuffle}
\end{table*}

\begin{table*}[t]
\centering
\small
\setlength{\tabcolsep}{6pt}
\begin{tabular}{llrrrr}
\hline
\textbf{Model} & \textbf{Domain} & \textbf{A} & \textbf{B} & \textbf{C} & \textbf{D} \\
\hline
Qwen3-0.6B & bioRxiv & 2,178 (99.8\%) & 3 (0.1\%) & 1 (0.0\%) & 1 (0.0\%) \\
Qwen3-0.6B & GloBI & 3,500 (100.0\%) & 0 & 0 & 0 \\
Qwen3-0.6B & Wikipedia & 1,927 (100.0\%) & 0 & 0 & 0 \\
\hline
SmolLM2-360M-Instruct & bioRxiv & 1 (0.0\%) & 2,168 (99.3\%) & 14 (0.6\%) & 0 \\
SmolLM2-360M-Instruct & GloBI & 138 (3.9\%) & 3,276 (93.6\%) & 85 (2.4\%) & 1 (0.0\%) \\
SmolLM2-360M-Instruct & Wikipedia & 85 (4.4\%) & 1,517 (78.7\%) & 323 (16.8\%) & 2 (0.1\%) \\
\hline
Qwen3-8B & bioRxiv & 731 (33.5\%) & 1,056 (48.4\%) & 319 (14.6\%) & 77 (3.5\%) \\
Qwen3-8B & GloBI & 1,640 (46.9\%) & 681 (19.5\%) & 757 (21.6\%) & 422 (12.1\%) \\
Qwen3-8B & Wikipedia & 432 (22.4\%) & 989 (51.3\%) & 398 (20.7\%) & 108 (5.6\%) \\
\hline
Llama 3.1-8B Instruct & bioRxiv & 716 (32.8\%) & 1,065 (48.8\%) & 331 (15.2\%) & 71 (3.3\%) \\
Llama 3.1-8B Instruct & GloBI & 1,655 (47.3\%) & 812 (23.2\%) & 829 (23.7\%) & 204 (5.8\%) \\
Llama 3.1-8B Instruct & Wikipedia & 453 (23.5\%) & 1,016 (52.7\%) & 373 (19.4\%) & 85 (4.4\%) \\

\hline
\end{tabular}
\caption{Answer letters emitted by four open models on the \emph{unshuffled} option sets (seed~0, greedy). Small models collapse to a single letter; larger models are more varied but still skewed on Wikipedia.}
\label{tab:appendix-answer-position-models}
\end{table*}

\section{Manual review notes on GloBI}
\label{app:globi_manual_review}
During validation design we inspected a small sample of GloBI items to identify ambiguity patterns that the audit should catch, such as underspecified hosts, mixed taxonomic ranks among options, or stems that could support more than one interaction label. These observations motivated the expert- and GPT-5.5--guided filtering described in Section~\ref{sec:human_validation}; the main evaluation uses only the validated subset. The easy/medium/hard strata elsewhere in the paper reflect frontier-model agreement rather than independent biological adjudication of every item.
\end{document}

%% file: tables/benchmark_examples_main.tex
\begin{figure*}[t]
\centering

\begin{minipage}[t]{0.38\textwidth}
\vspace{0pt}
\begin{examplebox}{bioRxiv}
\footnotesize
\textbf{Dimension:} Result Interpretation

\textbf{Level:} easy

\textbf{Question:} In a study of the sea anemone \textit{Nematostella vectensis}, researchers found that animals kept in environments with gravel substrate produced significantly more clonal progeny through transverse fission compared to those without substrate. Given that substrate enhances fission rates, what can be inferred about its role in asexual reproduction?

\textbf{Options:} A. Substrate provides a mechanical advantage that facilitates the physical process of fissioning. B. Substrate increases the genetic diversity of the clones produced during fission. C. Substrate reduces the metabolic waste that inhibits fission in high-density populations. D. Substrate alters the hormonal balance in \textit{Nematostella}, promoting faster growth.

\textbf{Gold:} A. Substrate provides a mechanical advantage that facilitates the physical process of fissioning.
\end{examplebox}
\end{minipage}
\hfill
\begin{minipage}[t]{0.58\textwidth}
\vspace{0pt}
\begin{examplebox}{Wikipedia}
\footnotesize
\textbf{Dimension:} Communication

\textbf{Level:} medium

\textbf{Question:} What is one of the distinct call types of the Andean flamingo?

\textbf{Options:} A. Whistle B. Chirp C. Growl D. Peep

\textbf{Gold:} D. Peep
\end{examplebox}

\vspace{0.8em}

\begin{examplebox}{GloBI}
\footnotesize
\textbf{Dimension:} Masked participant identification

\textbf{Level:} hard

\textbf{Question:} In an ecological interaction where a specific organism negatively affects the health of another, which taxon is likely the one being impacted if the first organism is identified as \textit{Parvioris mortoni}?

\textbf{Options:} A. \textit{Echinometra lucunter} B. \textit{Parvioris mortoni} C. \textit{Acanthaster planci} D. \textit{Archaster typicus}

\textbf{Gold:} D. \textit{Archaster typicus}
\end{examplebox}
\end{minipage}

\caption{Representative example items from each BAGEL source domain. \textbf{Level} indicates the reference difficulty label from Table~\ref{tab:bagel_difficulty_three_frontier} (agreement among GPT-5.4, Claude Opus~4.6, and Gemini 3.1 Flash Lite). The three panels are chosen to include \textbf{easy}, \textbf{medium}, and \textbf{hard} strata across domains.}
\label{fig:benchmark_examples_main}
\end{figure*}

%% file: tables/benchmark_examples_appendix.tex
\section{Example Benchmark Items}
\label{app:example_items}
\subsection{Wikipedia}
\begin{examplebox}{Example item}
\textbf{Dimension:} Taxonomy\\
\textbf{Level:} easy\\
\textbf{Question:} What is the scientific name of the Andean flamingo?\\
\textbf{Options:}\\
A. Phoenicopterus andinus\\
B. Phoenicoparrus andinus\\
C. Phoenicopterus chilensis\\
D. Phoenicoparrus chilensis\\
\textbf{Gold answer:} Phoenicoparrus andinus
\end{examplebox}
\begin{examplebox}{Example item}
\textbf{Dimension:} Geographic Distribution\\
\textbf{Level:} easy\\
\textbf{Question:} Where is the Andean flamingo native to?\\
\textbf{Options:}\\
A. The Amazon rainforest\\
B. The Andes mountains of South America\\
C. The coastal regions of Chile\\
D. The plains of Argentina\\
\textbf{Gold answer:} The Andes mountains of South America
\end{examplebox}
\begin{examplebox}{Example item}
\textbf{Dimension:} Diet\\
\textbf{Level:} easy\\
\textbf{Question:} What is the primary feeding strategy of the Andean flamingo?\\
\textbf{Options:}\\
A. Carnivorous predation\\
B. Filter feeding\\
C. Scavenging\\
D. Grazing on grass\\
\textbf{Gold answer:} Filter feeding
\end{examplebox}
\begin{examplebox}{Example item}
\textbf{Dimension:} Behavior\\
\textbf{Level:} easy\\
\textbf{Question:} How do Andean flamingos adapt their foraging behavior when grouped with other flamingo species?\\
\textbf{Options:}\\
A. They forage alone regardless of group\\
B. They adopt the foraging patterns of the species they are grouped with\\
C. They only forage at night\\
D. They stop foraging altogether\\
\textbf{Gold answer:} They adopt the foraging patterns of the species they are grouped with
\end{examplebox}
\begin{examplebox}{Example item}
\textbf{Dimension:} Communication\\
\textbf{Level:} medium\\
\textbf{Question:} What is one of the distinct call types of the Andean flamingo?\\
\textbf{Options:}\\
A. Chirp\\
B. Peep\\
C. Whistle\\
D. Growl\\
\textbf{Gold answer:} Peep
\end{examplebox}
\begin{examplebox}{Example item}
\textbf{Dimension:} Morphology\\
\textbf{Level:} easy\\
\textbf{Question:} What distinguishes the Andean flamingo from other flamingo species?\\
\textbf{Options:}\\
A. Its bright red plumage\\
B. Its yellow legs and three-toed feet\\
C. Its long neck\\
D. Its ability to fly at high altitudes\\
\textbf{Gold answer:} Its yellow legs and three-toed feet
\end{examplebox}
\begin{examplebox}{Example item}
\textbf{Dimension:} Habitat\\
\textbf{Level:} easy\\
\textbf{Question:} In which type of environment do Andean flamingos primarily live during the summer?\\
\textbf{Options:}\\
A. Forests\\
B. Salt lakes\\
C. Grasslands\\
D. Deserts\\
\textbf{Gold answer:} Salt lakes
\end{examplebox}
\begin{examplebox}{Example item}
\textbf{Dimension:} Cognition\\
\textbf{Level:} easy\\
\textbf{Question:} What cognitive ability is noted in kea regarding problem-solving?\\
\textbf{Options:}\\
A. They can memorize songs\\
B. They can solve logical puzzles\\
C. They can mimic human speech\\
D. They can navigate using stars\\
\textbf{Gold answer:} They can solve logical puzzles
\end{examplebox}

\subsection{GloBI}
\begin{examplebox}{Example item}
\textbf{Dimension:} Masked participant identification\\
\textbf{Level:} hard\\
\textbf{Question:} Which taxon is likely to be the host in the interaction where Pearsonema plica is an endoparasite?\\
\textbf{Options:}\\
A. Procyon lotor\\
B. Ursus americanus\\
C. Canis lupus\\
D. Lynx rufus\\
\textbf{Gold answer:} Procyon lotor\\
\textit{(Verbatim benchmark item; underlying GloBI record: Pearsonema plica endoparasiteof Procyon lotor.)}
\end{examplebox}
\begin{examplebox}{Example item}
\textbf{Dimension:} Masked interaction type inference\\
\textbf{Level:} easy\\
\textbf{Question:} In an ecological interaction where a tree species is negatively affected by a fungal organism, what type of interaction is most likely occurring?\\
\textbf{Options:}\\
A. The tree is providing nutrients to the fungus.\\
B. The fungus is benefiting at the expense of the tree.\\
C. The tree is benefiting the fungus without any harm.\\
D. The tree and fungus are mutually benefiting each other.\\
\textbf{Gold answer:} The fungus is benefiting at the expense of the tree.
\end{examplebox}

\subsection{bioRxiv}
\begin{examplebox}{Example item}
\textbf{Dimension:} Result Interpretation\\
\textbf{Level:} easy\\
\textbf{Question:} In a study of the sea anemone Nematostella vectensis, researchers found that animals kept in environments with gravel substrate produced significantly more clonal progeny through transverse fission compared to those without substrate. Given that the presence of substrate is shown to enhance fission rates, what can be inferred about the role of substrate in the asexual reproduction of Nematostella vectensis?\\
\textbf{Options:}\\
A. Substrate provides a mechanical advantage that facilitates the physical process of fissioning.\\
B. Substrate increases the genetic diversity of the clones produced during fission.\\
C. Substrate reduces the metabolic waste that inhibits fission in high-density populations.\\
D. Substrate alters the hormonal balance in Nematostella, promoting faster growth.\\
\textbf{Gold answer:} Substrate provides a mechanical advantage that facilitates the physical process of fissioning.
\end{examplebox}

%% file: main.bbl
\begin{thebibliography}{30}
\providecommand{\natexlab}[1]{#1}

\bibitem[{Abdin et~al.(2024)Abdin, Aneja, Behl, Bubeck, Eldan, Gunasekar,
  Harrison, Hewett, Javaheripi, Kauffmann et~al.}]{abdin2024phi4}
Marah Abdin, Jyoti Aneja, Harkirat Behl, S{\'e}bastien Bubeck, Ronen Eldan,
  Suriya Gunasekar, Michael Harrison, Russell~J. Hewett, Mojan Javaheripi,
  Piero Kauffmann, et~al. 2024.
\newblock Phi-4 technical report.
\newblock \emph{arXiv preprint arXiv:2412.08905}.

\bibitem[{Ben~Allal et~al.(2025)Ben~Allal, Lozhkov, Bakouch,
  Mart{\'\i}n~Bl{\'a}zquez, Penedo, Tunstall, Marafioti, Kydl{\'\i}{\v c}ek,
  Piqueres~Lajar{\'\i}n, Srivastav et~al.}]{benallal2025smollm2}
Loubna Ben~Allal, Anton Lozhkov, Elie Bakouch, Gabriel
  Mart{\'\i}n~Bl{\'a}zquez, Guilherme Penedo, Lewis Tunstall, Andr{\'e}s
  Marafioti, Hynek Kydl{\'\i}{\v c}ek, Agust{\'\i}n Piqueres~Lajar{\'\i}n,
  Vaibhav Srivastav, et~al. 2025.
\newblock Smollm2: When smol goes big -- data-centric training of a small
  language model.
\newblock \emph{arXiv preprint arXiv:2502.02737}.

\bibitem[{Bi et~al.(2023)Bi, Zhang, Xue, Ou, Ji, Zheng, and
  Chen}]{bi2023oceangpt}
Zhen Bi, Ningyu Zhang, Yida Xue, Yixin Ou, Daxiong Ji, Guozhou Zheng, and
  Huajun Chen. 2023.
\newblock Oceangpt: A large language model for ocean science tasks.
\newblock \emph{arXiv preprint arXiv:2310.02031}.

\bibitem[{Deng et~al.(2023)Deng, Zhang, He, Xu, Chen, Shi, Fu, Zhang, Wang,
  Zhou, Lin, and He}]{deng2023k2}
Cheng Deng, Tianhang Zhang, Zhongmou He, Yi~Xu, Qiyuan Chen, Yuanyuan Shi,
  Luoyi Fu, Weinan Zhang, Xinbing Wang, Chenghu Zhou, Zhouhan Lin, and Junxian
  He. 2023.
\newblock K2: A foundation language model for geoscience knowledge
  understanding and utilization.
\newblock \emph{arXiv preprint arXiv:2306.05064}.

\bibitem[{Dorm et~al.(2025)Dorm, Millard, Purves, Harfoot, and
  Mac~Aodha}]{dorm2025ecologicalknowledge}
Filip Dorm, Joseph Millard, Drew Purves, Michael Harfoot, and Oisin Mac~Aodha.
  2025.
\newblock Large language models possess some ecological knowledge, but how
  much?
\newblock \emph{bioRxiv}.

\bibitem[{{Gemma Team}(2024)}]{gemmateam2024gemma}
{Gemma Team}. 2024.
\newblock Gemma: Open models based on gemini research and technology.
\newblock \emph{arXiv preprint arXiv:2403.08295}.

\bibitem[{Gougherty and Clipp(2024)}]{gougherty2024ecological}
Andrew~V. Gougherty and Hannah~L. Clipp. 2024.
\newblock Testing the reliability of an ai-based large language model to
  extract ecological information from the scientific literature.
\newblock \emph{npj Biodiversity}, 3:13.

\bibitem[{Grattafiori et~al.(2024)Grattafiori, Dubey, Jauhri, Pandey, Kadian,
  Al-Dahle, Letman, Mathur, Schelten, Vaughan et~al.}]{grattafiori2024llama3}
Aaron Grattafiori, Abhimanyu Dubey, Abhinav Jauhri, Abhinav Pandey, Abhishek
  Kadian, Ahmad Al-Dahle, Aiesha Letman, Akhil Mathur, Alan Schelten, Alex
  Vaughan, et~al. 2024.
\newblock The llama 3 herd of models.
\newblock \emph{arXiv preprint arXiv:2407.21783}.

\bibitem[{Gu et~al.(2021)Gu, Tinn, Cheng, Lucas, Usuyama, Liu, Naumann, Gao,
  and Poon}]{gu2021domain}
Yu~Gu, Robert Tinn, Hao Cheng, Michael Lucas, Naoto Usuyama, Xiaodong Liu,
  Tristan Naumann, Jianfeng Gao, and Hoifung Poon. 2021.
\newblock Domain-specific language model pretraining for biomedical natural
  language processing.
\newblock \emph{ACM Transactions on Computing for Healthcare}, 3(1):1--23.

\bibitem[{Guo et~al.(2025)Guo, Li, and Xu}]{guo2025elle}
Jing Guo, Nan Li, and Ming Xu. 2025.
\newblock Environmental large language model evaluation (elle) dataset: A
  benchmark for evaluating generative ai applications in eco-environment
  domain.
\newblock \emph{arXiv preprint arXiv:2501.06277}.

\bibitem[{Hagiwara et~al.(2022)Hagiwara, Hoffman, Liu, Cusimano, Effenberger,
  and Zacarian}]{hagiwara2022beans}
Masato Hagiwara, Benjamin Hoffman, Jen-Yu Liu, Maddie Cusimano, Felix
  Effenberger, and Katie Zacarian. 2022.
\newblock Beans: The benchmark of animal sounds.
\newblock \emph{arXiv preprint arXiv:2210.12300}.

\bibitem[{Hagiwara et~al.(2024)Hagiwara, Miron, and Liu}]{hagiwara2024ispa}
Masato Hagiwara, Marius Miron, and Jen-Yu Liu. 2024.
\newblock Ispa: Inter-species phonetic alphabet for transcribing animal sounds.
\newblock \emph{arXiv preprint arXiv:2402.03269}.

\bibitem[{Hendrycks et~al.(2021)Hendrycks, Burns, Basart, Zou, Mazeika, Song,
  and Steinhardt}]{hendrycks2021mmlu}
Dan Hendrycks, Collin Burns, Steven Basart, Andy Zou, Mantas Mazeika, Dawn
  Song, and Jacob Steinhardt. 2021.
\newblock Measuring massive multitask language understanding.
\newblock In \emph{International Conference on Learning Representations
  (ICLR)}.

\bibitem[{Huang et~al.(2024)Huang, Guo, Guo, Tao, Lv, Sun, and
  Zhao}]{huang2024enviroexam}
Yu~Huang, Liang Guo, Wanqian Guo, Zhe Tao, Yang Lv, Zhihao Sun, and Dongfang
  Zhao. 2024.
\newblock Enviroexam: Benchmarking environmental science knowledge of large
  language models.
\newblock \emph{arXiv preprint arXiv:2405.11265}.

\bibitem[{Jiang et~al.(2023)Jiang, Sablayrolles, Mensch, Bamford, Chaplot,
  de~las Casas, Bressand, Lengyel, Lample, Saulnier et~al.}]{jiang2023mistral}
Albert~Q. Jiang, Alexandre Sablayrolles, Arthur Mensch, Chris Bamford,
  Devendra~Singh Chaplot, Diego de~las Casas, Florian Bressand, Gianna Lengyel,
  Guillaume Lample, Lucile Saulnier, et~al. 2023.
\newblock Mistral 7b.
\newblock \emph{arXiv preprint arXiv:2310.06825}.

\bibitem[{Jin et~al.(2019)Jin, Dhingra, Liu, Cohen, and Lu}]{jin2019pubmedqa}
Qiao Jin, Bhuwan Dhingra, Zhengping Liu, William~W. Cohen, and Xinghua Lu.
  2019.
\newblock Pubmedqa: A dataset for biomedical research question answering.
\newblock In \emph{Proceedings of the 2019 Conference on Empirical Methods in
  Natural Language Processing (EMNLP-IJCNLP)}, pages 2567--2577.

\bibitem[{Keck et~al.(2025)Keck, Broadbent, and Altermatt}]{keck2025massive}
Fran{\c c}ois Keck, Henry Broadbent, and Florian Altermatt. 2025.
\newblock Extracting massive ecological data on state and interactions of
  species using large language models.
\newblock \emph{bioRxiv}.

\bibitem[{Lewis et~al.(2020)Lewis, Perez, Piktus, Petroni, Karpukhin, Goyal,
  K{\"u}ttler, Lewis, Yih, Rockt{\"a}schel, Riedel, and Kiela}]{lewis2020rag}
Patrick Lewis, Ethan Perez, Aleksandra Piktus, Fabio Petroni, Vladimir
  Karpukhin, Naman Goyal, Heinrich K{\"u}ttler, Mike Lewis, Wen-tau Yih, Tim
  Rockt{\"a}schel, Sebastian Riedel, and Douwe Kiela. 2020.
\newblock Retrieval-augmented generation for knowledge-intensive {NLP} tasks.
\newblock In \emph{Advances in Neural Information Processing Systems
  (NeurIPS)}.

\bibitem[{Lu et~al.(2022)Lu, Mishra, Xia, Qiu, Chang, Zhu, Tafjord, Clark, and
  Kalyan}]{lu2022learn}
Pan Lu, Swaroop Mishra, Tony Xia, Liang Qiu, Kai-Wei Chang, Song-Chun Zhu,
  Oyvind Tafjord, Peter Clark, and Ashwin Kalyan. 2022.
\newblock Learn to explain: Multimodal reasoning via thought chains for science
  question answering.
\newblock In \emph{Advances in Neural Information Processing Systems
  (NeurIPS)}.

\bibitem[{Mallen et~al.(2023)Mallen, Asai, Zhong, Das, Khashabi, and
  Hajishirzi}]{mallen2023nottrust}
Alex Mallen, Akari Asai, Victor Zhong, Rajarshi Das, Daniel Khashabi, and
  Hannaneh Hajishirzi. 2023.
\newblock When not to trust language models: Investigating effectiveness of
  parametric and non-parametric memories.
\newblock In \emph{Proceedings of the 61st Annual Meeting of the Association
  for Computational Linguistics (ACL)}.

\bibitem[{Nentidis et~al.(2023)Nentidis, Katsimpras, Krithara, Lima~L{\'o}pez,
  Farr{\'e}-Maduell, Gasco, Krallinger, and Paliouras}]{nentidis2023bioasq}
Anastasios Nentidis, Georgios Katsimpras, Anastasia Krithara, Salvador
  Lima~L{\'o}pez, Eul{\`a}lia Farr{\'e}-Maduell, Luis Gasco, Martin Krallinger,
  and Georgios Paliouras. 2023.
\newblock Overview of bioasq 2023: The eleventh bioasq challenge on large-scale
  biomedical semantic indexing and question answering.
\newblock In \emph{Experimental IR Meets Multilinguality, Multimodality, and
  Interaction}, Lecture Notes in Computer Science.

\bibitem[{{OpenAI}(2023)}]{openai2023gpt4}
{OpenAI}. 2023.
\newblock Gpt-4 technical report.
\newblock \emph{arXiv preprint arXiv:2303.08774}.

\bibitem[{Ouyang et~al.(2022)Ouyang, Wu, Jiang, Almeida, Wainwright, Mishkin,
  Zhang, Agarwal, Slama, Ray, Schulman, Hilton, Kelton, Miller, Simens, Askell,
  Welinder, Christiano, Leike, and Lowe}]{ouyang2022instructgpt}
Long Ouyang, Jeff Wu, Xu~Jiang, Diogo Almeida, Carroll~L. Wainwright, Pamela
  Mishkin, Chong Zhang, Sandhini Agarwal, Katarina Slama, Alex Ray, John
  Schulman, Jacob Hilton, Fraser Kelton, Luke Miller, Maddie Simens, Amanda
  Askell, Peter Welinder, Paul Christiano, Jan Leike, and Ryan Lowe. 2022.
\newblock Training language models to follow instructions with human feedback.
\newblock In \emph{Advances in Neural Information Processing Systems
  (NeurIPS)}.

\bibitem[{Pezeshkpour and Hruschka(2024)}]{pezeshkpour2024large}
Pouya Pezeshkpour and Estevam Hruschka. 2024.
\newblock Large language models sensitivity to the order of options in
  multiple-choice questions.
\newblock In \emph{Findings of the Association for Computational Linguistics:
  NAACL 2024}, pages 2006--2017.

\bibitem[{Poelen et~al.(2014)Poelen, Simons, and Mungall}]{poelen2014globi}
Jorrit~H. Poelen, James~D. Simons, and Chris~J. Mungall. 2014.
\newblock Global biotic interactions: An open infrastructure to share and
  analyze species-interaction datasets.
\newblock \emph{Ecological Informatics}, 24:148--159.

\bibitem[{Robinson et~al.(2024)Robinson, Miron, Hagiwara, Weck, Keen, Alizadeh,
  Narula, Geist, and Pietquin}]{robinson2024naturelmaudio}
David Robinson, Marius Miron, Masato Hagiwara, Benno Weck, Sara Keen, Milad
  Alizadeh, Gagan Narula, Matthieu Geist, and Olivier Pietquin. 2024.
\newblock Naturelm-audio: an audio-language foundation model for bioacoustics.
\newblock \emph{arXiv preprint arXiv:2411.07186}.

\bibitem[{Stevens et~al.(2024)Stevens, Wu, Thompson, Campolongo, Song, Carlyn,
  Dong, Dahdul, Stewart, Berger-Wolf, Chao, and Su}]{stevens2024bioclip}
Samuel Stevens, Jiaman Wu, Matthew~J. Thompson, Elizabeth~G. Campolongo,
  Chan~Hee Song, David~Edward Carlyn, Li~Dong, Wasila~M. Dahdul, Charles
  Stewart, Tanya Berger-Wolf, Wei-Lun Chao, and Yu~Su. 2024.
\newblock Bioclip: A vision foundation model for the tree of life.
\newblock In \emph{Proceedings of the IEEE/CVF Conference on Computer Vision
  and Pattern Recognition (CVPR)}.

\bibitem[{Team et~al.(2025)Team, Kamath, Ferret, Pathak, Vieillard, Merhej,
  Perrin, Matejovicova, Ramé, Rivière, Rouillard, Mesnard, Cideron, bastien
  Grill, Ramos, Yvinec, Casbon, Pot, Penchev, Liu, Visin, Kenealy, Beyer, Zhai,
  Tsitsulin, Busa-Fekete, Feng, Sachdeva, Coleman, Gao, Mustafa, Barr,
  Parisotto, Tian, Eyal, Cherry, Peter, Sinopalnikov, Bhupatiraju, Agarwal,
  Kazemi, Malkin, Kumar, Vilar, Brusilovsky, Luo, Steiner, Friesen, Sharma,
  Sharma, Gilady, Goedeckemeyer, Saade, Feng, Kolesnikov, Bendebury, Abdagic,
  Vadi, György, Pinto, Das, Bapna, Miech, Yang, Paterson, Shenoy, Chakrabarti,
  Piot, Wu, Shahriari, Petrini, Chen, Lan, Choquette-Choo, Carey, Brick,
  Deutsch, Eisenbud, Cattle, Cheng, Paparas, Sreepathihalli, Reid, Tran, Zelle,
  Noland, Huizenga, Kharitonov, Liu, Amirkhanyan, Cameron, Hashemi,
  Klimczak-Plucińska, Singh, Mehta, Lehri, Hazimeh, Ballantyne, Szpektor,
  Nardini, Pouget-Abadie, Chan, Stanton, Wieting, Lai, Orbay, Fernandez,
  Newlan, yeong Ji, Singh, Black, Yu, Hui, Vodrahalli, Greff, Qiu, Valentine,
  Coelho, Ritter, Hoffman, Watson, Chaturvedi, Moynihan, Ma, Babar, Noy, Byrd,
  Roy, Momchev, Chauhan, Sachdeva, Bunyan, Botarda, Caron, Rubenstein,
  Culliton, Schmid, Sessa, Xu, Stanczyk, Tafti, Shivanna, Wu, Pan, Rokni,
  Willoughby, Vallu, Mullins, Jerome, Smoot, Girgin, Iqbal, Reddy, Sheth,
  Põder, Bhatnagar, Panyam, Eiger, Zhang, Liu, Yacovone, Liechty, Kalra, Evci,
  Misra, Roseberry, Feinberg, Kolesnikov, Han, Kwon, Chen, Chow, Zhu, Wei,
  Egyed, Cotruta, Giang, Kirk, Rao, Black, Babar, Lo, Moreira, Martins,
  Sanseviero, Gonzalez, Gleicher, Warkentin, Mirrokni, Senter, Collins, Barral,
  Ghahramani, Hadsell, Matias, Sculley, Petrov, Fiedel, Shazeer, Vinyals, Dean,
  Hassabis, Kavukcuoglu, Farabet, Buchatskaya, Alayrac, Anil, Dmitry, Lepikhin,
  Borgeaud, Bachem, Joulin, Andreev, Hardin, Dadashi, and
  Hussenot}]{gemmateam2025gemma3technicalreport}
Gemma Team, Aishwarya Kamath, Johan Ferret, Shreya Pathak, Nino Vieillard,
  Ramona Merhej, Sarah Perrin, Tatiana Matejovicova, Alexandre Ramé, Morgane
  Rivière, Louis Rouillard, Thomas Mesnard, Geoffrey Cideron, Jean bastien
  Grill, Sabela Ramos, Edouard Yvinec, Michelle Casbon, Etienne Pot, Ivo
  Penchev, and 197 others. 2025.
\newblock \href {https://arxiv.org/abs/2503.19786} {Gemma 3 technical report}.
\newblock \emph{Preprint}, arXiv:2503.19786.

\bibitem[{Yang et~al.(2025)Yang, Li, Yang, Zhang, Hui, Zheng, Yu, Gao, Huang,
  Lv et~al.}]{yang2025qwen3}
An~Yang, Anfeng Li, Baosong Yang, Beichen Zhang, Binyuan Hui, Bo~Zheng, Bowen
  Yu, Chang Gao, Chengen Huang, Chenxu Lv, et~al. 2025.
\newblock Qwen3 technical report.
\newblock \emph{arXiv preprint arXiv:2505.09388}.

\bibitem[{Zheng et~al.(2024)Zheng, Zhou, Meng, Zhou, and Huang}]{zheng2023llms}
Chujie Zheng, Hao Zhou, Fandong Meng, Jie Zhou, and Minlie Huang. 2024.
\newblock Large language models are not robust multiple choice selectors.
\newblock \emph{arXiv preprint arXiv:2309.03882}.

\end{thebibliography}
